\documentclass[11pt]{article}

\usepackage[margin=1in]{geometry}
\usepackage{amsmath,amssymb,amsthm,mathtools}
\usepackage{array}
\usepackage{booktabs}
\usepackage{enumitem}
\usepackage{graphicx}
\usepackage{hyperref}

\hypersetup{
  colorlinks=true,
  linkcolor=blue,
  citecolor=blue,
  urlcolor=blue
}

\newtheorem{theorem}{Theorem}[section]
\newtheorem{lemma}[theorem]{Lemma}
\newtheorem{proposition}[theorem]{Proposition}
\newtheorem{corollary}[theorem]{Corollary}
\theoremstyle{definition}
\newtheorem{definition}[theorem]{Definition}
\newtheorem{remark}[theorem]{Remark}

\newcommand{\R}{\mathbb{R}}

\newcommand{\E}{\mathbb{E}}
\newcommand{\Prob}{\Pr}
\newcommand{\calH}{\mathcal{H}}
\newcommand{\calC}{\mathcal{C}}
\newcommand{\calR}{\mathcal{R}}
\newcommand{\calE}{\mathcal{E}}
\newcommand{\calG}{\mathcal{G}}
\newcommand{\TopK}{\operatorname{TopK}}
\newcommand{\Emb}{\operatorname{Emb}}
\newcommand{\Out}{\operatorname{Out}}
\newcommand{\ReLU}{\operatorname{ReLU}}
\newcommand{\Unif}{\operatorname{Unif}}
\newcommand{\Andreev}{\mathrm{Andreev}}
\newcommand{\CDT}{\mathrm{CDT}}
\newcommand{\bin}{\mathrm{bin}}
\newcommand{\ind}[1]{\mathbf{1}\!\left[#1\right]}

\makeatletter
\renewcommand{\@fnsymbol}[1]{\ensuremath{\ifcase#1\or \dagger\or \ddagger\or \mathsection\or \mathparagraph\or \|\or **\or \dagger\dagger\or \ddagger\ddagger \else\@ctrerr\fi}}
\makeatother

\title{\vspace{-2.8em}Average-Case Lower Bounds For Sparse-Attention LLMs\\(Or: Why Open-Source LLMs Are Falling Behind?)}
\author{Youlong Ding\thanks{The Hebrew University of Jerusalem, Israel. Email: \texttt{youlong.ding@mail.huji.ac.il}.}}
\date{}

\begin{document}
\maketitle
\vspace{-2.5em}

\begin{center}
\begin{minipage}{\textwidth}
\centering
\makeatletter
\def\@captype{figure}
\makeatother
\includegraphics[width=0.90\textwidth]{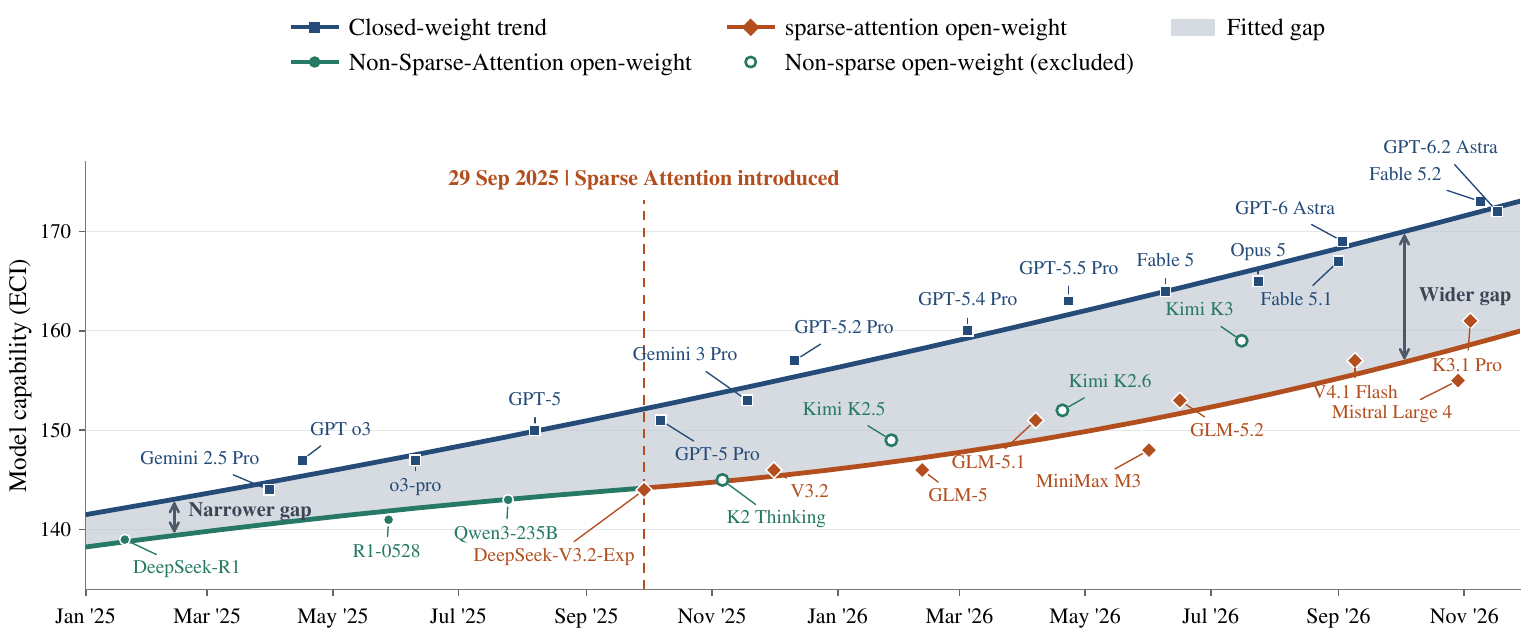}
\vspace{-0.5em}
\caption{The widening capability gap between closed-weight LLMs and open-weight LLMs adopting sparse attention. The vertical dashed line marks the introduction of DeepSeek Sparse Attention (DSA) in September~2025.}
\label{fig:capability-gap}
\end{minipage}
\end{center}
\vspace{-0.8em}

\begin{abstract}
Since DeepSeek-V3.2 introduced DeepSeek Sparse Attention (DSA) in September~2025, dynamic Top-$k$ sparse attention has swept across the open-weight LLM ecosystem (DeepSeek through V4.1, GLM, MiniMax, Qwen, Hunyuan, and Kimi) for its long-context efficiency. Yet since mid-2026, open-source frontier progress has visibly plateaued, opening a widening capability gap behind leading closed-source models.

We show that sparse attention imposes a fundamental representational bottleneck. We formulate a general model of multi-layer dynamic Top-$k$ sparse attention, dubbed the \emph{Sparse Transformer}, allowing arbitrary scoring and hidden-state updates (subsuming DSA, blockwise indexers, sliding-window hybrids, softmax attention, etc). On Andreev's function over $n$ uniformly random bits (which is computable with $100\%$ accuracy by a $2$-layer Full Attention Transformer of dimension $O(\log n)$), we prove an unconditional \emph{average-case} lower bound for sparse attention. In particular, for any constant depth $L \ge 2$, polylogarithmic hidden-state bit capacity, and $O(1)$ indexers per layer, no $L$-layer Sparse Transformer achieves accuracy better than $\frac{1}{2} + n^{-\Omega(1)}$ even at sparsity $k = n^{\alpha}$ for any constant $\alpha < \frac{1}{L-1}$.

Technically, our proof combines Andreev's random-restriction compression framework with a \emph{Top-$k$ Switching Lemma} generalizing H\aa stad's Switching Lemma and a final-layer constant-key pruning argument.
\end{abstract}

\section{Introduction}
\label{sec:intro}

In September~2025, DeepSeek introduced \emph{DeepSeek Sparse
Attention} (DSA) with the release of DeepSeek-V3.2-Exp (and
subsequently DeepSeek-V3.2~\cite{deepseek2025v32}), marking a
departure from the dense softmax self-attention that had been
a cornerstone of the Transformer architecture since its
introduction~\cite{vaswani2017attention}.
To alleviate the quadratic $O(n^2)$ prefill complexity and
linear per-token KV-cache traffic of full attention, DSA
employs a lightweight \emph{Lightning Indexer}
to dynamically select only a small subset of past tokens
(such as $k = 2048$ out of a $128\text{K}$-token context)
for each query.\footnote{Strictly speaking, DSA retains an $O(n^2)$ complexity, since the indexer still examines all preceding
tokens. The savings come from replacing expensive dense
attention with lightweight dense indexing followed by
sparse attention.}
The immediate engineering gains were substantial: DSA
reportedly cut long-context training and serving costs by
more than half while maintaining comparable performance
to its dense predecessor on standard short- and
long-context evaluations.

What began as an efficiency experiment rapidly solidified into consensus dogma across the open-source ecosystem. DeepSeek retained and progressively sharpened sparse attention in every subsequent flagship release, from V3.2 through DeepSeek-V4 and DeepSeek-V4.1 (the latter pushing sparse indexing even further via Compressed Sparse Attention~2 and cross-layer index reuse). Almost every other major open-weight developer soon followed the same trajectory, converging on dynamic Top-$k$ sparse attention and sparse-hybrid variants (Table~\ref{tab:adoption}): Zhipu AI adopted DSA in GLM-5~\cite{glm5_2026} (and cross-layer index sharing in subsequent iterations), MiniMax introduced blockwise Top-$k$ MiniMax Sparse Attention (MSA)~\cite{minimax2026msa} in MiniMax-M3, Alibaba's Qwen introduced four-token micro-block Qwen Sparse Attention (QSA) in Qwen3.8-Flash-Next and Qwen4-Max, Tencent's Hunyuan (HY) deployed gated DSA with cross-layer index caching in Hy4, and Moonshot AI (Kimi) replaced full attention with gated DSA in Kimi~K3.1~Pro. By mid-2026, sparse attention had acquired the aura of an indisputable architectural truth, as though full quadratic attention had merely been a wasteful redundancy waiting to be pruned away.

\begin{table}[t]
\centering
\small
\renewcommand{\arraystretch}{1.22}
\begin{tabular}{@{}>{\raggedright\arraybackslash}p{0.17\textwidth}>{\raggedright\arraybackslash}p{0.26\textwidth}>{\raggedright\arraybackslash}p{0.27\textwidth}>{\raggedright\arraybackslash}p{0.21\textwidth}@{}}
\toprule
\textbf{Lab / Family} & \textbf{Sparse Mechanism} & \textbf{Representative Models} & \textbf{Earliest Adoption} \\
\midrule
DeepSeek & DSA (Lightning Indexer); CSA2 (index reuse) & DeepSeek-V3.2-Exp, V3.2, V4, V4-Pro, V4.1-Flash & Sep 2025 \newline (V3.2-Exp) \\
Zhipu AI (GLM) & DSA; cross-layer IndexShare & GLM-5, GLM-5.1, GLM-5.2, GLM-5.3 / 5.4 & Feb 2026 \newline (GLM-5) \\
MiniMax & MSA (blockwise Top-$k$ over GQA) & MiniMax-M3, M3.1-Flash & Jun 2026 \newline (MiniMax-M3) \\
Alibaba (Qwen) & QSA (4-token micro-block Top-$k$ indexer) & Qwen3.8-Flash-Next, Qwen4-Max & Aug 2026 \newline (Qwen3.8-Flash-Next) \\
Tencent (HY) & Gated DSA with cross-layer IndexCache & Hy4 (Preview \& Full), Hy4-Flash & Aug 2026 \newline (Hy4 Preview) \\
Moonshot (Kimi) & Hybrid KDA + Gated DSA & Kimi K3.1 Pro & Oct 2026 \newline (Kimi K3.1 Pro) \\
\bottomrule
\end{tabular}
\caption{Widespread adoption of dynamic Top-$k$ sparse attention across major open-weight LLM families (2025--2026). All mechanisms listed above rely on pairwise query-to-key (or query-to-block) Top-$k$ selection and are subsumed by Definition~\ref{def:sparse-transformer}.}
\label{tab:adoption}
\end{table}

Yet beginning around mid-2026, an unexpected divergence became hard to ignore. After years in which open-weight models steadily closed the distance to proprietary systems, open-source frontier capabilities visibly slowed, and a widening performance gap re-opened relative to the strongest closed-source models. Independent tracking by Epoch AI reflects this widening gap: whereas their October~2025 analysis found that open-weight models lagged the closed-weight state of the art by roughly 3~months on average~\cite{epoch2025lag}, their May~2026 follow-up on the Epoch Capabilities Index (ECI) reported that the lag had grown to 4~months~\cite{epoch2026lag}; by late~2026, the fitted ECI trend for sparse-attention open-weight models trails the closed-weight frontier by roughly 6~months (Figure~\ref{fig:capability-gap}). Notably, the onset of this divergence coincides strikingly with the open-weight industry's wholesale transition to sparse attention. Popular post-mortems have pointed to compute asymmetries, reinforcement-learning data pipelines, or post-training recipes. While those factors certainly play a role, this paper investigates a more basic, structural question: 
\begin{center}
    \emph{Does the widening capability gap between open-weight and proprietary LLMs reflect a fundamental limitation of the architectures adopted by open-weight models? \footnote{The author has been skeptical of sparse attention as a foundational scaling direction ever since DeepSeek first introduced DSA in V3.2. Seeing nearly every major open-weight lab subsequently adopt sparse attention—and witnessing open-source model performance visibly stagnate behind the closed-source frontier—convinced the author that it is necessary to articulate and prove this limitation from first principles.}}
\end{center}

\paragraph{Beyond naive receptive-field arguments.}
A fundamental challenge in establishing theoretical limitations
of sparse attention is that \emph{multi-layer sparse attention
is not locally blind}. With a sparsity budget of $k = 2048$
tokens per layer, even $L = 2$ layers can yield a receptive
field of $k^2 > 4 \times 10^6$ tokens, far exceeding a
$128\text{K}$ or $1\text{M}$ context window. In other words,
modern sparse-attention LLMs operate deep inside the
\emph{supercritical regime} $k^L \gg n$, where even a fixed
$k$-ary communication tree can span the entire sequence and
compute global functions such as $\mathrm{MAJORITY}_n$ and
$\mathrm{PARITY}_n$. A meaningful lower bound must therefore
go beyond such naive receptive-field arguments and establish
a separation within the supercritical regime.

Furthermore, unlike in classical complexity theory, where
a worst-case lower bound often suffices to establish a
separation between two computational models, in the context
of LLMs we are arguably more interested in their
\emph{average-case} performance. A worst-case separation
is consistent with sparse attention matching full attention
on almost all inputs, failing only on a negligible fraction
of them. We therefore seek a substantially stronger
separation: an explicit task that full attention solves
perfectly, but on which sparse attention cannot outperform
random guessing by more than a small advantage
$\varepsilon$ on uniformly random inputs.

\subsection{Our Contributions}

We establish an unconditional average-case separation between
full attention and multi-layer dynamic Top-$k$ sparse attention,
even in the supercritical regime $k^L > n$.

\begin{itemize}[leftmargin=1.5em,itemsep=4pt]

  \item \textbf{A general model of sparse attention
  (Section~\ref{sec:model}).}
  We introduce the
  \emph{$(L,k,H_{\mathrm{idx}},B)$-Sparse Transformer},
  an abstraction of multi-layer dynamic Top-$k$ sparse
  attention. One natural and key modeling choice is to require the
  indexers to score tokens in a \emph{pairwise} manner\footnote{This prevents the selection mechanism itself from encoding
  arbitrary computations over the entire input.}, while allowing
  arbitrary scoring functions and hidden-state updates.
  Our model subsumes DeepSeek Sparse
  Attention~\cite{deepseek2025v32} and accommodates
  blockwise sparse attention mechanisms, such as
  MoBA~\cite{lu2025moba} and
  MSA~\cite{minimax2026msa}, with appropriate adjustments
  to the model parameters.
  In particular, our lower bound does not rely on any
  specific form of attention scoring, value aggregation,
  or feed-forward computation.

  \item \textbf{An unconditional average-case lower bound
  (Theorem~\ref{thm:main}).}
  We study Andreev's function
  $\Andreev_N(T,x)=T_{S(x)}$ over $n=\Theta(N)$ input bits,
  which requires computing an address $S(x)$ through
  global parity aggregations and then retrieving the
  corresponding entry from a truth table $T$.
  While a $2$-layer causal full-attention Transformer
  with model dimension $d=O(\log N)$ computes this
  function with $100\%$ accuracy
  (Proposition~\ref{prop:full-upper}), we show that any
  $L$-layer Sparse Transformer achieving accuracy
  $\frac12+\varepsilon$ on uniformly random inputs
  must satisfy
  \[
    \Big(C\,H_{\mathrm{idx}}
    \big(k+\log_2(NH_{\mathrm{idx}})\big)\Big)^{L-1}
    \cdot (B+\log_2 N)
    =\Omega\!\left(
      \frac{\varepsilon^2 N}{\log N\log\log N}
    \right),
  \]
  for some absolute constant $C$.

  In particular, for any constant depth $L\ge 2$,
  $H_{\mathrm{idx}}=O(1)$, and
  $B=\mathrm{polylog}(N)$, sparse attention achieves
  accuracy at most $\frac12+N^{-\Omega(1)}$ even at
  sparsity $k=N^\alpha$ for any constant
  \[
    \frac{1}{L}<\alpha<\frac{1}{L-1}.
  \]
  Thus, sparse attention fails to compute the function
  on average even when its receptive field is
  polynomially larger than the entire input.

  \item \textbf{Matching sparse upper bounds
  (Proposition~\ref{prop:sparse-upper}).}
  We complement our lower bound with explicit
  constructions of Sparse Transformers that compute
  Andreev's function exactly.
  Specifically, a causal $(L,k,1,B)$-Sparse Transformer
  with $B\ge\log_2 N$ achieves $100\%$ accuracy whenever
  either
  \[
    k^{L-1}\ge N,
  \]
  or
  \[
    k^L\ge 4N
    \qquad\text{and}\qquad
    kB\ge 4N,
  \]
  with $k\ge 4$ in the latter case.
  In particular, for constant depth and polylogarithmic
  hidden-state capacity, our lower and upper bounds
  match up to polylogarithmic factors in the sparsity
  budget $k$.

  \item \textbf{A Top-$k$ Switching Lemma
  (Lemma~\ref{lem:topk-switching}).}
  As an additional technical contribution that may be
  of independent interest, we establish a generalization
  of H\aa stad's Switching Lemma~\cite{hastad1986}.
  Whereas the canonical decision tree in the classical
  switching lemma searches for the first satisfied term
  of an ordered DNF, our generalization considers the
  task of identifying the \emph{first $k$ satisfied terms}.
  We show that, for any ordered list of conjunctions
  of width at most $W$, the canonical decision tree
  for this task collapses under a random restriction
  $\rho\sim\calR_p$ to depth less than $r$, except with
  probability at most
  \[
    2^k\left(\frac{16pW}{1-p}\right)^r.
  \]
  Our proof extends Razborov's encoding
  argument~\cite{razborov1995bounded,beame1994switching}.
  This lemma serves as the main technical tool for
  analyzing input-dependent Top-$k$ selection across
  multiple layers.

\end{itemize}

Finally, we note that the main message of this work is conceptual rather than
quantitative: sparse attention may come at a fundamental
barrier.
Our goal is to put this perspective on a formal footing,
by introducing a computational model that captures the
essential features of modern sparse attention architectures
and establishing a first set of meaningful theoretical
separations from full attention.
Rather than seeking the sharpest possible bounds within
this model, we view this work as initiating a systematic
study of the computational limitations of sparse attention.

\subsection{Implications for Long-Context Reasoning}
\label{subsec:llm-implications}

\paragraph{Global aggregation followed by retrieval.}
Andreev's function captures a fundamental computational
primitive in long-context reasoning: \emph{global information
aggregation followed by content-addressed retrieval}.
Consider, for example, debugging a large code repository.
A model may need to first analyze function calls and data
dependencies scattered across multiple files to identify
the function responsible for a bug, and then retrieve and
examine the implementation of that function. The key
observation is that \emph{what to retrieve} is itself the
result of a computation over the context.
Andreev's function abstracts precisely this computational
pattern: the model must first aggregate globally dispersed
information to determine the address $S(x)$, and then
retrieve the corresponding entry $T_{S(x)}$ from the
context.

\paragraph{Reachability is not computation.}
Full attention realizes this computational primitive in
just two layers: the first layer aggregates information
from the entire context to compute $S(x)$, and the second
layer performs the content-addressed retrieval.
One might expect multi-layer sparse attention to achieve
the same once its receptive field covers the entire context,
i.e., when $k^L \ge n$.
Our result shows that this intuition is misleading.
Even when $k^L \gg n$, sparse attention can fail to realize
this primitive, achieving essentially no better than random
guessing on uniformly random inputs.
The bottleneck is not whether information from every token
can eventually reach the output, but whether the model can
aggregate and retrieve information in a coordinated manner
within the available computational depth.
This reveals a fundamental distinction between full and
sparse attention that cannot be captured by receptive-field
analysis alone.

\subsection{Related Work}

\paragraph{Dynamic sparse attention in frontier LLMs.}
Early sparse attention architectures relied on static patterns (sliding windows, dilated blocks, global sink tokens), which struggled on arbitrary long-range retrieval. Following earlier research proposals for trainable blockwise routing such as Mixture of Block Attention (MoBA)~\cite{lu2025moba}, open-weight frontier models rapidly shifted toward \emph{content-dependent, trainable} Top-$k$ sparsity beginning with DeepSeek-V3.2's fine-grained DeepSeek Sparse Attention (DSA) via the Lightning Indexer~\cite{deepseek2025v32} (later extended with cross-layer index reuse in V4/V4.1): Zhipu AI adopted DSA in GLM-5~\cite{glm5_2026}, MiniMax proposed blockwise MiniMax Sparse Attention (MSA)~\cite{minimax2026msa} in MiniMax-M3, Alibaba introduced micro-block Qwen Sparse Attention (QSA) in Qwen3.8-Flash-Next/Qwen4, Tencent's Hunyuan deployed gated DSA with index caching in Hy4, and Moonshot AI integrated gated DSA into Kimi~K3.1~Pro. All of these mechanisms are subsumed by Definition~\ref{def:sparse-transformer}.

\paragraph{Theoretical limitations of Transformers.}
A growing line of work analyzes the representational strengths and limitations of self-attention using communication complexity and circuit complexity. Sanford, Hsu, and Telgarsky~\cite{sanford2023representational} separated self-attention from RNNs and MLPs via sparse averaging and established communication-complexity lower bounds on triple detection. Peng, Narayanan, and Papadimitriou~\cite{peng2024limitations} proved fundamental limitations of single-layer Transformers on function composition via communication complexity, linking them to hallucinations and compositional failures. Most recently, Chen, Peng, and Wu~\cite{chen2025limitations} established the first unconditional lower bounds against \emph{multi-layer} decoder-only Transformers: introducing a multi-party autoregressive communication model and an indistinguishable-decomposition technique, they proved that any constant-$L$-layer causal Transformer requires polynomial dimension $n^{\Omega(1)}$ to sequentially compose $L$ functions, yielding sharp depth--width trade-offs, an encoder--decoder separation, and a provable advantage for chain-of-thought reasoning. Whereas~\cite{chen2025limitations} exploits causal masking to separate depths of \emph{full} attention, our work addresses a complementary and timely architectural question: unconditionally separating $2$-layer causal \emph{Full Attention} from (even bidirectional) $L$-layer \emph{dynamic Top-$k$ Sparse Attention} in the average case, strictly inside the supercritical regime $k^L > n$.

\paragraph{Boolean circuit complexity foundations.}
Our proof builds directly upon classical tools in Boolean circuit complexity: Andreev's function and random-restriction shrinkage method~\cite{andreev1987}, H\aa stad's Switching Lemma~\cite{hastad1986} and Razborov's encoding proof~\cite{razborov1995bounded,beame1994switching}, Multi-Switching Lemmas~\cite{imp2012,hastad2014}, and the average-case information-theoretic Andreev compression framework used by Kane and Williams~\cite{kane2016superlinear} alongside depth-$L$ threshold circuit trade-offs~\cite{ips1997}. We spell out the exact relationship between our proof and these standard complexity-theoretic tools in Section~\ref{subsec:complexity-connections}.

\section{Formal Model and Main Results}
\label{sec:model}

\subsection{Multi-Layer Top-\texorpdfstring{$k$}{k} Sparse Attention}

A central subtlety in formalizing sparse attention lies in
capturing the computational power of the Top-$k$ selection
mechanism. On the one hand, restricting attention to fixed,
input-independent patterns would fail to capture the dynamic
nature of modern sparse attention architectures. On the other
hand, allowing the selected subset $S_t^{(\ell)} \subseteq [n]$
to be an arbitrary function of the entire input
$z \in \{0,1\}^n$ would make the model overly powerful:
even with $L=1$ and $k=1$, the selection mechanism could
encode the output of any Boolean function $f(z)$ into the
identity of the selected token.

To strike a balance between these two extremes, we require
the Top-$k$ selection to be based on pairwise scores between
query and key hidden states, with fixed, input-independent
tie-breaking. Beyond this restriction, we allow both the
scoring functions and the hidden-state updates to be
arbitrary. As we discuss below, this abstraction captures
the essential features of modern dynamic sparse attention
architectures while remaining sufficiently general that
our lower bounds are independent of their specific
architectural choices.

\begin{definition}[$(L, k, H_{\mathrm{idx}}, B)$-Sparse Transformer]
\label{def:sparse-transformer}
Let $n \ge 1$ be the sequence length, $L \ge 1$ the number of layers, $k \ge 1$ the sparsity budget, $H_{\mathrm{idx}} \ge 1$ the number of independent Top-$k$ indexers per layer, and $B \ge 1$ the hidden-state bit capacity. Let $\calH = \{0, 1\}^B$. For each position $t \in [n]$, fix a nonempty candidate set $\calC_t \subseteq [n]$ (with $\calC_t = [t]$ in the causal/decoder-only setting and $\calC_t = [n]$ in the bidirectional/encoder-only setting) and a fixed, input-independent strict total order $\prec_t$ on $[n]$ for tie-breaking.

An \emph{$(L, k, H_{\mathrm{idx}}, B)$-Sparse Transformer} computes a function $g : \{0, 1\}^n \to \{0, 1\}$ by generating hidden states $h_t^{(\ell)} \in \calH$ for $t \in [n]$ and $\ell \in \{0, 1, \dots, L\}$ as follows:
\begin{enumerate}[leftmargin=2em]
  \item \textbf{Input layer ($\ell = 0$):} For each position $t \in [n]$,
  \[
    h_t^{(0)} = \Emb_t(z_t) \in \calH,
  \]
  where $\Emb_t : \{0, 1\} \to \calH$ depends only on the single input symbol $z_t$ and position $t$.

  \item \textbf{Sparse attention layer ($\ell \in \{1, \dots, L\}$):} For each position $t \in [n]$:
  \begin{enumerate}[label=(\alph*),leftmargin=1.5em]
    \item \emph{Pairwise indexing:} For each indexer head $r \in [H_{\mathrm{idx}}]$ and each candidate key $j \in \calC_t$, compute the pairwise score
    \[
      I_{t, j}^{(\ell, r)} = \phi_{t, j}^{(\ell, r)}\!\left(h_t^{(\ell-1)},\, h_j^{(\ell-1)}\right) \in \R,
    \]
    where $\phi_{t, j}^{(\ell, r)} : \calH \times \calH \to \R$ is an arbitrary pairwise function.

    \item \emph{Top-$k$ selection:} Define the strict total order $\succ_t$ on $\R \times [n]$ by $(a, j) \succ_t (a', j')$ if and only if $a > a'$, or ($a = a'$ and $j \prec_t j'$). Let
    \[
      S_t^{(\ell, r)} = \TopK_{j \in \calC_t}\!\left(I_{t, j}^{(\ell, r)},\, k;\, \prec_t\right) \subseteq \calC_t
    \]
    be the unique subset of $\calC_t$ of size $\min(k, |\calC_t|)$ consisting of the indices whose pairs $\big(I_{t, j}^{(\ell, r)}, j\big)$ are largest under $\succ_t$. Set $S_t^{(\ell)} = \bigcup_{r=1}^{H_{\mathrm{idx}}} S_t^{(\ell, r)}$.

    \item \emph{Sparse aggregation and update:} Update the hidden state via an arbitrary function $\psi_t^{(\ell)}$:
    \[
      h_t^{(\ell)} = \psi_t^{(\ell)}\!\left(h_t^{(\ell-1)},\; \Big\{\big(j, h_j^{(\ell-1)}\big) : j \in S_t^{(\ell)}\Big\}\right) \in \calH.
    \]
  \end{enumerate}

  \item \textbf{Readout:} Output $g(z) = \Out\!\left(h_n^{(L)}\right) \in \{0, 1\}$ for an arbitrary map $\Out : \calH \to \{0, 1\}$.
\end{enumerate}
\end{definition}

\begin{remark}[Relation to DSA and Modern Sparse-Attention Architectures]
\label{rem:dsa-relation}
Definition~\ref{def:sparse-transformer} strictly generalizes real-world dynamic sparse attention architectures:
\begin{itemize}[leftmargin=1.5em,itemsep=2pt]
  \item \emph{DeepSeek Sparse Attention (DSA):} In DeepSeek-V3.2~\cite{deepseek2025v32} (and its adoption in GLM-5~\cite{glm5_2026} and Hunyuan Hy4), the Lightning Indexer computes pairwise scores
  \[
    I_{t, j} \;=\; \sum_{u=1}^{H^I} w_{t, u}^I\, \ReLU\!\left(\langle q_{t, u}^I, k_j^I \rangle\right)
  \]
  and feeds a \emph{single} shared Top-$k$ index set $S_t^{(\ell)}$ to all Multi-head Latent Attention (MLA) heads. This corresponds directly to the special case $H_{\mathrm{idx}} = 1$.
  \item \emph{Blockwise sparse attention (MoBA, MSA, QSA):} Selecting the Top-$k_{\mathrm{blk}}$ blocks (or micro-blocks) of $b$ tokens each using a query-to-block pairwise score (where each block's summary key can be computed in one preliminary sparse layer of budget $b$) is subsumed by setting $k = b \cdot k_{\mathrm{blk}}$ (with $H_{\mathrm{idx}}$ equal to the number of GQA groups in MSA~\cite{minimax2026msa}, or $b = 4$ in QSA).
  \item \emph{Positional encodings, normalization, and MLPs:} Allowing $\phi_{t, j}^{(\ell, r)}$ and $\psi_t^{(\ell)}$ to depend arbitrarily on $(t, j)$ subsumes RoPE, ALiBi, and arbitrary learned positional encodings. Allowing arbitrary $\psi_t^{(\ell)}$ subsumes multi-head softmax attention over $S_t^{(\ell)}$, MLA, LayerNorm/RMSNorm, residual connections, and arbitrary-width signed MLPs.
  \item \emph{Sliding-window hybrids and attention sinks:} Adding a local sliding window of width $w_{\mathrm{win}}$ or $s_{\mathrm{sink}}$ fixed sink tokens is simply a positional Top-$(w_{\mathrm{win}} + s_{\mathrm{sink}})$ indexer head, which at most replaces $k$ by $k + w_{\mathrm{win}} + s_{\mathrm{sink}}$ (or increments $H_{\mathrm{idx}}$ by $1$).
  \item \emph{Monotone score normalizations:} Any order-preserving transformation applied across $j$ before Top-$k$ (such as softmax or affine rescaling of scores) leaves $S_t^{(\ell, r)}$ unchanged.
  \item \emph{Bidirectional vs.\ causal:} Our lower bound holds for arbitrary candidate sets $\calC_t \subseteq [n]$ (including bidirectional attention $\calC_t = [n]$), while our Full Attention upper bounds are strictly causal ($\calC_t = [t]$).
\end{itemize}
\end{remark}

\subsection{Andreev's Function and Main Separation}
\label{sec:results}

To separate Full Attention from $L$-layer Sparse Attention in the supercritical regime $k^L > n$, we use Andreev's function~\cite{andreev1987,kane2016superlinear}.

\begin{definition}[Andreev's Function]
\label{def:andreev}
Let $N = 2^m$ with integer $m = \log_2 N \ge 2$, let $s = \lfloor N / m \rfloor \ge 1$, and let $n = N + m s$ (so $N < n \le 2N$). Partition an input sequence $z = (T, x) \in \{0, 1\}^n$ into:
\begin{itemize}[leftmargin=1.5em,itemsep=2pt]
  \item a \emph{truth table} $T = (T_u)_{u \in \{0,1\}^m} \in \{0, 1\}^N$ occupying positions $1, \dots, N$ (identifying $j \in [N]$ with its $m$-bit binary expansion $\bin(j) \in \{0, 1\}^m$); and
  \item $m$ \emph{address blocks} $x = (x^{(1)}, \dots, x^{(m)}) \in (\{0, 1\}^s)^m$ occupying positions $N+1, \dots, n$, where $x^{(r)} = (x_{r, 1}, \dots, x_{r, s})$ occupies block $G_r \subseteq \{N+1, \dots, n\}$.
\end{itemize}
Define the address $S(x) = \big(S_1(x), \dots, S_m(x)\big) \in \{0, 1\}^m$ by $S_r(x) = \bigoplus_{i=1}^s x_{r, i}$ for each $r \in [m]$. \emph{Andreev's function} $\Andreev_N : \{0, 1\}^n \to \{0, 1\}$ is
\[
  \Andreev_N(T, x) \;=\; T_{S(x)}.
\]
\end{definition}

We first record the upper bounds showing that Full Attention solves $\Andreev_N$ effortlessly in depth~$2$ (or depth~$3$ with logarithmic MLP width), and characterizing exactly when a Pairwise Sparse Transformer can solve it (full constructions are given in Appendix~\ref{app:upper-bounds}).

\begin{proposition}[Full Attention Upper Bounds]
\label{prop:full-upper}
$\Andreev_N(T, x)$ is computed with $100\%$ worst-case accuracy by:
\begin{enumerate}[label=(\alph*),leftmargin=2em,itemsep=2pt]
  \item a $2$-layer causal Full Softmax Attention Transformer with $H = 1$ head, model dimension $d = O(\log N)$, $O(\log N)$-bit precision ($B = O(\log^2 N)$), and a $2$-layer ReLU MLP of width $O(N)$; and
  \item a $3$-layer causal Full Softmax Attention Transformer with $H = O(\log N)$ heads, model dimension $d = O(\log N)$, $O(\log N)$-bit precision, and a $2$-layer ReLU MLP of width $O(\log N)$.
\end{enumerate}
\end{proposition}

\begin{proposition}[Matching Sparse Attention Upper Bounds]
\label{prop:sparse-upper}
For any $L \ge 2$, a causal $(L, k, 1, B)$-Pairwise Sparse Transformer computes $\Andreev_N(T, x)$ with $100\%$ worst-case accuracy whenever either:
\begin{enumerate}[label=(\alph*),leftmargin=2em,itemsep=2pt]
  \item $k^{L-1} \ge N$ and $B \ge \log_2 N$; or
  \item $k \ge 4$, $k^L \ge 4 N$, $k B \ge 4 N$, and $B \ge \log_2 N$.
\end{enumerate}
\end{proposition}

Our main theorem establishes that surpassing random guessing on $\Andreev_N$ requires the product of the $(L-1)$-layer sparse fan-in and the hidden-state bit capacity to reach $\tilde{\Omega}(\varepsilon^2 N)$—showing that Proposition~\ref{prop:sparse-upper}(a) is essentially tight for bounded-capacity models even in the average case.

\begin{theorem}[Average-Case Lower Bound for $L$-Layer Pairwise Sparse Attention]
\label{thm:main}
There exists an absolute constant $C \ge 1$ such that the following holds. For any integer $L \ge 2$, $N = 2^m$ ($m \ge 2$), and $n = N + m\lfloor N/m \rfloor$, if an $(L, k, H_{\mathrm{idx}}, B)$-Pairwise Sparse Transformer $g : \{0, 1\}^n \to \{0, 1\}$ satisfies
\[
  \Prob_{(T, x) \sim \Unif(\{0,1\}^n)}\!\Big[g(T, x) = \Andreev_N(T, x)\Big] \;\ge\; \frac{1}{2} + \varepsilon
\]
for some $\varepsilon \in (0, \frac{1}{2}]$, then
\[
  \Big(C\, H_{\mathrm{idx}}\big(k + \log_2(N H_{\mathrm{idx}})\big)\Big)^{L-1} \cdot (B + \log_2 N) \;=\; \Omega\!\left(\frac{\varepsilon^2 N}{\log N \log\log N}\right).
\]
\end{theorem}

The following corollary highlights the two sharpest operational consequences of Theorem~\ref{thm:main}: failure throughout the supercritical regime $k^{L-1} \ll N < k^L$, and an explicit depth overhead relative to Full Attention.

\begin{corollary}[Supercritical Separation and Depth Penalty]
\label{cor:supercritical}
Let $c, C' > 0$ be suitable absolute constants.
\begin{enumerate}[label=(\alph*),leftmargin=2em,itemsep=2pt]
  \item \emph{Supercritical failure at fixed depth:} Fix any constant depth $L \ge 2$, constant $H_{\mathrm{idx}} = O(1)$, hidden-state capacity $B = \mathrm{polylog}(N)$, and sparsity $k = N^{\alpha}$ with constant $\alpha \in (\frac{1}{L}, \frac{1}{L - 1})$. Then $k^L = N^{\alpha L} \gg N$ (so an $L$-layer Sparse Transformer can already aggregate all $n$ tokens via a $k$-ary tree and solve $\mathrm{MAJORITY}_n$ and $\mathrm{PARITY}_n$), yet every $(L, k, H_{\mathrm{idx}}, B)$-Pairwise Sparse Transformer achieves average-case accuracy at most $\frac{1}{2} + \tilde{O}\!\big(N^{-(1 - \alpha(L-1))/2}\big) = \frac{1}{2} + N^{-\Omega(1)}$ on $\Andreev_N$. In particular, for $L = 2$ and $H_{\mathrm{idx}} = 1$, sparse attention achieves accuracy at most $\frac{1}{2} + \tilde{O}(N^{-\delta/2})$ even at $k = N^{1 - \delta}$ (deep inside $k^2 = N^{2 - 2\delta} \gg N$).
  \item \emph{Depth penalty relative to Full Attention:} Achieving average-case accuracy $\frac{1}{2} + \varepsilon$ on $\Andreev_N$ requires depth
  \[
    L \;\ge\; 1 + \frac{\log_2\!\left(\frac{c\, \varepsilon^2 N}{(B + \log_2 N)\log_2 N \log_2\log_2 N}\right)}{\log_2\!\big(C'\, H_{\mathrm{idx}}(k + \log_2(N H_{\mathrm{idx}}))\big)},
  \]
  whereas Full Softmax Attention achieves $100\%$ worst-case accuracy at depth $L = 2$ (or $L = 3$ with $O(\log N)$-width MLPs).
\end{enumerate}
\end{corollary}

\begin{remark}[Complexity Barrier for $L \ge 4$ with $k^2 \gg N$]
\label{rem:tc0-barrier}
One may naturally ask whether an unconditional lower bound in Definition~\ref{def:sparse-transformer} can simultaneously achieve $L \ge 4$ and $k^2 \gg N$ (for example, $L = 4$ and $k = N^{0.6}$). Observe that with $B = O(\log N)$ and arbitrary positional scoring $\phi_{t,j}$, an $(L, k, 1, B)$-Pairwise Sparse Transformer can simulate any depth-$L$ Majority circuit with $O(L n)$ gates and $O(L n k)$ wires. For $L \ge 4$ and $k = N^{0.6}$, proving a lower bound against depth-$4$ Majority circuits of $O(n^{1.6})$ wires for \emph{any} explicit function in $\mathsf{NP}$ (let alone one in Full Attention) is a long-standing open problem in circuit complexity: Impagliazzo, Paturi, and Saks~\cite{ips1997} reach $n^{1 + c^{-L}}$ wires (still within $k^L \lesssim n$), and Kane and Williams~\cite{kane2016superlinear} reached depth~$2$ and a restricted depth~$3$ via Andreev's function. Thus, pushing past $k^{L-1} < N < k^L$ to $k^2 \gg N$ at arbitrary depth $L$ without additional architectural restrictions (such as non-negative value mixing or symmetry constraints) hits the depth-$L$ $\mathsf{TC}^0$ wire barrier.
\end{remark}

\section{Technical Overview}
\label{sec:overview}

Before presenting the full formal proofs, we explain the proof of Theorem~\ref{thm:main} at an intuitive level and clarify its precise relationship to standard techniques in Boolean circuit complexity.

\subsection{High-Level Proof Intuition}
\label{subsec:intuition}

\paragraph{The chicken-and-egg trap of $\Andreev_N$.}
View the input $z = (T, x)$ as a long context split into two halves: an incompressible knowledge base $T \in \{0,1\}^N$ ($N$ uniformly random bits), and a dispersed reasoning prompt $x$ partitioned into $m = \log_2 N$ blocks of size $s \approx N/m$, whose parities encode the target address $S(x) \in \{0,1\}^m$. A $2$-layer Full Attention Transformer resolves this cleanly because Layer~$1$ computes all $m$ block sums globally in parallel (recovering $S(x)$ at token $n$), and Layer~$2$ uses $S(x)$ as a content-addressed query to fetch $T_{S(x)}$.

A Sparse Attention Transformer with $k^{L-1} B \ll N < k^L$ is caught in a dilemma:
\begin{itemize}[leftmargin=1.5em,itemsep=2pt]
  \item \emph{It cannot defer the lookup by smuggling $T$ to the output.} The final token reads at most $H_{\mathrm{idx}} k$ states of $B$ bits each in Layer~$L$, so unless $H_{\mathrm{idx}} k B \gtrsim N$, it cannot simply gather the entire table $T$ into Layer~$L$ and index into it once $S(x)$ is ready.
  \item \emph{It cannot finish computing $S(x)$ before Layer~$L$.} If the attention graph were \emph{static} (fixed independent of the input), then by Layer~$L-1$ each token could depend on at most $k^{L-1} \ll N$ positions of $x$, so it would miss almost every block of $x$ and have no clue what $S(x)$ is.
\end{itemize}
The entire mathematical difficulty is that dynamic Top-$k$ attention is \emph{not} static: the attention graph itself adapts to $(T, x)$ by inspecting \emph{all} $n$ pairwise scores at every layer! Our proof shows that under a random restriction of the address bits $x$, \emph{dynamic pairwise Top-$k$ selection cannot outperform static $k$-ary fan-in.}

\paragraph{Step 1: Restricting $x$ while keeping $T$ intact (Andreev compression).}
Apply a random restriction $\rho \sim \calR_p$ strictly to the address positions $V_x$: each address bit $x_j$ is independently left alive ($*$) with tiny probability $p = \Theta(\frac{\log\log N}{N/\log N})$, and otherwise fixed to a uniform random bit in $\{0,1\}$. Crucially, the truth table $T$ is \emph{never restricted}. Because $p s = 2\ln(2m)$, with high probability every one of the $m$ address blocks retains at least one alive variable; pruning each block down to a \emph{single} alive variable $y_r \in \{0,1\}$ yields a pruned restriction $\tilde{\rho}$.

The beauty of Andreev's function is that under $\tilde{\rho}$, the restricted address is simply $S(\tilde{\rho} \circ y) = y \oplus c(\tilde{\rho})$ for a fixed shift $c(\tilde{\rho}) \in \{0,1\}^m$. Thus, the restricted truth table $y \mapsto \Andreev_N(T, \tilde{\rho} \circ y)$ is still a \emph{bijective permutation of all $N$ random bits of $T$}. By Fano's and Kraft's inequalities (Lemma~\ref{lem:fano}), if the Transformer $g$ predicts $\Andreev_N$ with accuracy $\frac{1}{2} + \varepsilon$, then any prefix-free description $E(\tilde{\rho}, T)$ of the restricted function $y \mapsto g(T, \tilde{\rho} \circ y)$ must have expected length at least $\Omega(\varepsilon^2 N)$ bits. Our goal reduces to showing that the restricted sparse network can be compressed into $\tilde{O}\!\big((C H_{\mathrm{idx}} (k + \log(N H_{\mathrm{idx}})))^{L-1} (B + \log N)\big)$ bits.

\paragraph{Step 2: Warm-up intuition at $L = 2$ (Why one sparse layer collapses).}
To see why dynamic Top-$k$ selection collapses under random restrictions, consider first a single sparse layer ($\ell = 1$) with $H_{\mathrm{idx}} = 1$. Fix $T$ and a query token $t$ (say $t \in [N]$, so its Layer-$0$ state is fixed). Each table key $j \in [N]$ has a fixed score, while each address key $j \in V_x$ has two possible scores corresponding to $x_j \in \{0,1\}$. Imagine listing all candidate $(j, b)$ events from highest score down to lowest, and revealing the random restriction $\rho_1 \sim \calR_{p_1}$ as we scan down this sorted list to determine the Top-$k$ winners $S_t^{(1)}$.

Whenever our scan encounters an address key $j$ whose bit is \emph{fixed} by $\rho_1$ to its high-scoring value—an event occurring with constant probability $(1 - p_1)/2$—that key permanently locks into one of the Top-$k$ slots. Once $k$ keys have permanently locked in, their scores establish an unbreakable floor: \emph{no candidate lower in the sorted list can ever enter $S_t^{(1)}$}, regardless of how the remaining alive ($*$) variables are eventually set! Consequently, the scan terminates after encountering only $O(k + \log n)$ alive variables (with high probability), so the entire Layer-$1$ hidden state $h_t^{(1)}|_{\rho_1}$ depends on at most $O(k + \log n)$ alive variables—it collapses to a shallow decision tree!

\paragraph{Step 3: General depth $L$ via the Top-$k$ Switching Lemma.}
How do we iterate this collapse across $L - 1$ layers? Decompose the overall restriction $\rho \sim \calR_p$ into $L$ independent stages:
\[
  \rho \;=\; \underbrace{\rho_1 \circ \rho_2 \circ \dots \circ \rho_{L-1}}_{\rho_{\mathrm{switch}}} \;\circ\; \rho_{\mathrm{final}}.
\]
Suppose inductively that after Layers $1, \dots, \ell - 1$ (under $\rho_1 \circ \dots \circ \rho_{\ell-1}$), every token's state $h_j^{(\ell-1)}$ is computed by a decision tree $D_j$ of depth at most $w$ over the surviving variables of $V_x$. Fix a query token $t$ and indexer head $r$ in Layer~$\ell$. For every candidate key $j \in \calC_t$, combine $D_t$ and $D_j$ into a single decision tree $D_{t,j}$ of depth at most $W = 2w$: each leaf $\lambda$ of $D_{t,j}$ corresponds to a conjunction $T_{j, \lambda}$ of width at most $W$ that \emph{uniquely determines} both states $(h_t^{(\ell-1)}, h_j^{(\ell-1)})$ and hence the exact pairwise score $I_{t, j}^{(\ell, r)}$.

Now collect \emph{all} leaves across \emph{all} candidate keys $j \in \calC_t$ into one master list $\mathcal{T}_{t,r} = (T_1, T_2, \dots, T_M)$, sorted in decreasing order of score $(I_{t,j}^{(\ell, r)}, j)$ under $\succ_t$. On every completion of the surviving variables, for each $j \in \calC_t$ \emph{exactly one} leaf of $D_{t,j}$ evaluates to~$1$. Therefore, determining the exact Top-$k$ set $S_t^{(\ell, r)}$—together with $h_t^{(\ell-1)}$ and all selected states—is completely equivalent to finding the \textbf{first $k$ conjunctions in $\mathcal{T}_{t,r}$ that evaluate to $1$}!

When $k = 1$, finding the first satisfied conjunction in an ordered list of width-$W$ conjunctions is literally the canonical decision tree of a width-$W$ DNF—the exact object handled by H\aa stad's Switching Lemma! By generalizing Razborov's encoding proof from $k = 1$ to arbitrary $k \ge 1$, we prove the \textbf{Top-$k$ Switching Lemma} (Lemma~\ref{lem:topk-switching}): under $\rho_\ell \sim \calR_{p_\ell}$, the canonical Top-$k$ decision tree $\CDT_k(\mathcal{T}_{t,r}; \rho_\ell)$ has depth $\ge r_0$ with probability at most $2^k \big(\frac{16 p_\ell W}{1 - p_\ell}\big)^{r_0}$. Setting $p_\ell = \frac{1}{160 w}$ (so $\frac{16 p_\ell W}{1 - p_\ell} < \frac{1}{4}$) and $r_0 = k + \lceil 10 \log_2(n H_{\mathrm{idx}}) \rceil$ drives the failure probability below $(n H_{\mathrm{idx}})^{-20}$, allowing a union bound over all $n$ tokens and $H_{\mathrm{idx}}$ indexers. Concatenating the $H_{\mathrm{idx}}$ trees per token preserves the inductive depth bound $w = H_{\mathrm{idx}} r_0$ through Layer $L - 1$ (Lemma~\ref{lem:multi-collapse}), at a cumulative survival probability of $p_{\mathrm{switch}} = \Theta(w^{-(L-2)})$.

\paragraph{Step 4: Stopping at Layer $L - 1$ (leaf shrinkage and constant-key domination).}
Crucially, we do \emph{not} switch Layer~$L$—doing so would spend another factor of $w$ and only rule out $k^L \ll N$, missing the supercritical regime! Instead, we apply the leftover restriction $\rho_{\mathrm{final}} \sim \calR_{p_{\mathrm{final}}}$ (with $p_{\mathrm{final}} = p / p_{\mathrm{switch}} = \Theta(w^{L-2} p)$) directly to the depth-$w$ Layer-$(L-1)$ decision trees. By a simple inductive recurrence (Lemma~\ref{lem:leaf-shrinkage}), a depth-$w$ decision tree hit by $\calR_q$ collapses completely to a \emph{constant} except with probability $O(w q)$, and its expected leaf count when non-constant is $O(w q)$. Summing over all $n$ tokens, the total expected number of leaves across all \emph{non-constant} Layer-$(L-1)$ tokens is only
\[
  O(n w p_{\mathrm{final}}) \;=\; O\!\big(160^{L-1} w^{L-1} n p\big) \;=\; O\!\big(160^{L-1} w^{L-1} \log N \log\log N\big).
\]
Finally, to encode the Layer-$L$ output $y \mapsto g(T, \tilde{\rho} \circ y)$ (Lemma~\ref{lem:encoding}), we emit:
\begin{enumerate}[leftmargin=2em,itemsep=2pt]
  \item the decision trees (with $B$-bit leaf labels) of every \emph{non-constant} Layer-$(L-1)$ token, costing $\tilde{O}(160^{L-1} w^{L-1}(B + \log N))$ expected bits; and
  \item the decision tree $D_n$ of the query token $n$, together with—for each leaf $\lambda$ of $D_n$ and each indexer head $r$—only the \textbf{Top-$k$ constant keys} in $\calC_n$ (costing $O(H_{\mathrm{idx}} k(B + \log N))$ expected bits).
\end{enumerate}
Why can we discard all other constant keys? Because whenever $h_n^{(L-1)}$ takes the value at leaf $\lambda$, every constant key outside those Top-$k$ constant keys is strictly beaten by all $k$ of them regardless of $y$, and therefore can \emph{never} enter $S_n^{(L, r)}$! Comparing the total expected description length against the Fano lower bound $\Omega(\varepsilon^2 N)$ yields Theorem~\ref{thm:main}. Table~\ref{tab:stages} summarizes the parameter evolution across the $L$ stages.

\begin{table}[t]
\centering
\small
\renewcommand{\arraystretch}{1.28}
\begin{tabular}{@{}lll>{\raggedright\arraybackslash}p{0.44\textwidth}@{}}
\toprule
\textbf{Stage} & \textbf{Restriction} & \textbf{Survival prob.} & \textbf{Effect on hidden states} \\
\midrule
Input ($\ell = 0$) & none & $1$ & Each $h_t^{(0)}$ is a decision tree of depth $\le 1$ \\
Layer $1$ & $\rho_1$ & $p_1 = 1/160$ & Top-$k$ Switching collapses all $h_t^{(1)}$ to depth $\le w$ \\
Layers $2 \dots L-1$ & $\rho_2 \circ \dots \circ \rho_{L-1}$ & $p_\ell = 1/(160w)$ & Inductively keeps all $h_t^{(\ell)}$ at depth $\le w$ \\
Final shrinkage & $\rho_{\mathrm{final}}$ & $p_{\mathrm{final}} = p / p_{\mathrm{switch}}$ & Only $\tilde{O}(160^{L-1} w^{L-1})$ non-constant leaves survive \\
Layer $L$ readout & pruning $\rho \to \tilde{\rho}$ & $1$ var/block & Constant-key pruning compresses $g|_{T,\tilde{\rho}}$ against the $\Omega(\varepsilon^2 N)$ Fano lower bound \\
\bottomrule
\end{tabular}
\caption{Summary of the $L$-stage random restriction schedule with $w = H_{\mathrm{idx}}\big(k + \lceil 10\log_2(n H_{\mathrm{idx}}) \rceil\big)$ and $p_{\mathrm{switch}} = 160^{-(L-1)} w^{-(L-2)}$.}
\label{tab:stages}
\end{table}

\subsection{Relationship to Classical Complexity Theory Techniques}
\label{subsec:complexity-connections}

A reader familiar with Boolean circuit complexity will rightly recognize that the core engine of our proof rests on relatively standard—and time-tested—techniques from circuit lower bounds, paired with two clean adaptations specific to dynamic sparse attention. We find it valuable to make this lineage completely explicit:

\begin{enumerate}[leftmargin=2em,itemsep=4pt]
  \item \textbf{Andreev's method and information-theoretic compression (standard).}
  Using Andreev's function $\Andreev_N(T, x)$ to amplify a random-restriction shrinkage bound into a lower bound by keeping the truth table $T$ unrestricted dates back to Andreev~\cite{andreev1987}. Casting the final counting step as an \emph{average-case} prefix-free compression lower bound via Kraft's and Fano's inequalities is likewise standard in modern average-case circuit complexity (e.g., Kane and Williams~\cite{kane2016superlinear}).

  \item \textbf{H\aa stad's Switching Lemma and Razborov's encoding argument (standard, generalized to Top-$k$).}
  Our Top-$k$ Switching Lemma (Lemma~\ref{lem:topk-switching}) is a direct generalization of Razborov's encoding proof~\cite{razborov1995bounded,beame1994switching} of H\aa stad's Switching Lemma~\cite{hastad1986}. In classical $\mathsf{AC}^0$ proofs, an $\mathrm{OR}$ gate over width-$W$ conjunctions stops at the \emph{first} ($k = 1$) satisfied conjunction. Extending the canonical decision tree to collect the \emph{first $k$} satisfied conjunctions changes Razborov's encoding in just one place: between consecutive queried blocks $T_{i_{m-1}}$ and $T_{i_m}$, the decoder must also know the count $h_m \ge 0$ of already-satisfied conjunctions skipped in between. Because $\sum_{m=1}^b h_m \le k - 1$, specifying $(h_1, \dots, h_r)$ costs only $\binom{r + k - 1}{k - 1} \le 2^{r+k}$ advice choices—multiplying the classic switching failure probability by a mere $2^k$ factor rather than a naive $\binom{M}{k}$ union bound. Note also how this differs from the \emph{Multi-Switching Lemma} of Impagliazzo--Matthews--Paturi~\cite{imp2012} and H\aa stad~\cite{hastad2014}: Multi-Switching collapses a \emph{parallel family} of independent DNFs into a shared partial decision tree, whereas Top-$k$ Switching tracks $k$ \emph{sequential} hits along a single priority-ordered list of conjunctions.

  \item \textbf{Adaptation~1: Serializing arbitrary pairwise Top-$k$ states into DNF lists.}
  What makes the reduction from Pairwise Sparse Transformers to Top-$k$ Switching work—despite $B$-bit hidden states, real-valued scores $\phi_{t,j}$, and arbitrary nonlinear updates $\psi_t$—is that decision-tree depth does not care about the output alphabet size $|\calH| = 2^B$! Once Layer-$(\ell-1)$ states are represented as depth-$w$ trees over $V_x$, every leaf of the joint tree $D_{t,j}$ fixes both states and thus produces a static score and a width-$2w$ conjunction. Sorting all such leaves into a single priority list strips away both $B$ and $\psi_t$ during Layers $1, \dots, L-1$, so the hidden-state bit capacity $B$ enters only \emph{linearly} at the final compression step rather than exponentially inside the switching lemma.

  \item \textbf{Adaptation~2: Stopping at Layer $L - 1$ to breach the supercritical barrier $k^L > N$.}
  In standard $\mathsf{AC}^0$ lower bounds, one switches every layer all the way to the output gate. Doing so here would lose a factor of $w$ at Layer~$L$ as well and only yield $k^L \gtrsim N$, which is vacuous in the supercritical regime $k^L > N$. Stopping the switching induction at Layer $L - 1$, shrinking almost all depth-$w$ trees to constants via the residual restriction $\rho_{\mathrm{final}}$ (Lemma~\ref{lem:leaf-shrinkage}), and exploiting the \emph{Top-$k$ domination property} of constant keys at Layer~$L$ (Lemma~\ref{lem:encoding}) is what saves a full factor of $k$ and unlocks the separation in $k^{L-1} B \ll N < k^L$.
\end{enumerate}

In short, the paper does not require esoteric new machinery: rather, it shows that once dynamic pairwise Top-$k$ sparse attention is viewed through the right abstraction, classical random-restriction and switching-lemma technology already exposes a sharp, unconditional average-case bottleneck—all the way up to the $\mathsf{TC}^0$ wire barrier of~\cite{ips1997,kane2016superlinear}.

\section{Compression-via-Restriction Framework}
\label{sec:framework}

We now formalize the compression-via-restriction framework for Theorem~\ref{thm:main}. Throughout Sections~\ref{sec:framework} and~\ref{sec:proof-main}, we assume $N = 2^m \ge 32$ ($m \ge 5$) so that $s = \lfloor N / m \rfloor \ge 6$ and the survival probability $p$ defined below satisfies $p \in (0, 1)$; for the finitely many values $N < 32$, Theorem~\ref{thm:main} holds trivially by choosing an appropriate hidden constant.

\subsection{Random and Pruned Andreev Restrictions}

Let $V_x = \{N+1, \dots, n\} = \bigsqcup_{r=1}^m G_r$ denote the $m s$ address positions. A \emph{restriction} on $V_x$ is a map $\rho : V_x \to \{0, 1, *\}$. Positions with $\rho(j) = *$ are called \emph{alive}; all other positions are \emph{fixed}. For $p \in (0, 1)$, let $\calR_p$ be the product distribution over restrictions $\rho : V_x \to \{0, 1, *\}$ that independently sets each $j \in V_x$ to $*$ with probability $p$, to $0$ with probability $\frac{1-p}{2}$, and to $1$ with probability $\frac{1-p}{2}$.

Set the target survival probability
\begin{equation}
\label{eq:p-def}
  p \;=\; \frac{2 \ln(2m)}{s},
\end{equation}
which satisfies $0 < p < 1$ for all $m \ge 5$ (since $2\ln(10) \approx 4.61 < 6 = \lfloor 32 / 5 \rfloor$, and $s / \ln(2m)$ is increasing for $m \ge 5$). Define the good event
\[
  \calE_{\mathrm{alive}} \;=\; \Big\{\rho : \forall r \in [m],\; \exists\, j \in G_r \text{ with } \rho(j) = *\Big\}.
\]
By a union bound over the $m$ blocks,
\begin{equation}
\label{eq:e-alive}
  \Prob_{\rho \sim \calR_p}[\neg \calE_{\mathrm{alive}}] \;\le\; m (1 - p)^s \;\le\; m e^{-p s} \;=\; \frac{1}{4m} \;\le\; \frac{1}{20} \qquad (m \ge 5),
\end{equation}
so in particular $\Prob[\calE_{\mathrm{alive}}] \ge \frac{1}{2}$.

Conditioned on $\calE_{\mathrm{alive}}$, define the \emph{pruned Andreev restriction} $\tilde{\rho}$ from $\rho$ as follows: independently for each block $G_r$ ($r \in [m]$), choose one surviving position $i_r^* \in \{j \in G_r : \rho(j) = *\}$ uniformly at random to keep alive ($\tilde{\rho}(i_r^*) = *$), and set $\tilde{\rho}(j) \sim \Unif(\{0, 1\})$ independently for all other $j \in G_r$ with $\rho(j) = *$. For any $y = (y_1, \dots, y_m) \in \{0, 1\}^m$, let $\tilde{\rho} \circ y \in \{0, 1\}^{ms}$ denote the completion setting $x_{i_r^*} = y_r$ for each $r \in [m]$.

\begin{lemma}[Exact Marginality under Pruning]
\label{lem:pruning}
Conditioned on $\calE_{\mathrm{alive}}$, if $y \sim \Unif(\{0, 1\}^m)$ is independent of $\tilde{\rho}$, then $x = \tilde{\rho} \circ y$ is distributed identically to $\Unif(\{0, 1\}^{ms})$. Moreover, there exists a vector $c(\tilde{\rho}) \in \{0, 1\}^m$ depending only on $\tilde{\rho}$ such that $S(\tilde{\rho} \circ y) = y \oplus c(\tilde{\rho})$ for all $y \in \{0, 1\}^m$.
\end{lemma}

\begin{proof}
The event $\calE_{\mathrm{alive}}$ depends only on the set of positions assigned $*$ by $\rho$, and not on any fixed bit values. Conditioned on $\calE_{\mathrm{alive}}$ and the set of alive positions of $\rho$, every coordinate $j \in V_x$ is independently uniform in $\{0, 1\}$ under $\tilde{\rho} \circ y$—whether fixed by $\rho$, fixed during pruning, or assigned by $y_r$. Thus $\tilde{\rho} \circ y \sim \Unif(\{0, 1\}^{ms})$. Finally, setting $c_r(\tilde{\rho}) = \bigoplus_{j \in G_r \setminus \{i_r^*\}} \tilde{\rho}(j)$ gives $S_r(\tilde{\rho} \circ y) = y_r \oplus c_r(\tilde{\rho})$ for every $r \in [m]$.
\end{proof}

\subsection{Information-Theoretic Lower Bound on Encoding Length}

\begin{lemma}[Compression Lower Bound via Fano's Inequality]
\label{lem:fano}
Let $g : \{0, 1\}^n \to \{0, 1\}$ satisfy $\Prob_{(T, x)}[g(T, x) = \Andreev_N(T, x)] \ge \frac{1}{2} + \varepsilon$ for $\varepsilon \in (0, \frac{1}{2}]$. Let $T \sim \Unif(\{0, 1\}^N)$ and let $\tilde{\rho}$ be drawn independently from the pruned distribution conditioned on $\calE_{\mathrm{alive}}$. Suppose $E \in \{0, 1\}^*$ is a (possibly randomized) binary string coupled with $(\tilde{\rho}, T)$ such that:
\begin{enumerate}[label=(\roman*),leftmargin=2em,itemsep=2pt]
  \item for every fixed value $\tilde{\rho}_0$ of $\tilde{\rho}$, the support of $E$ given $\tilde{\rho} = \tilde{\rho}_0$ is a prefix-free set of binary strings; and
  \item from $(\tilde{\rho}, E)$ (together with the fixed network $g$, but without $T$), one can deterministically reconstruct the entire restricted function $y \mapsto g(T, \tilde{\rho} \circ y)$ for all $y \in \{0, 1\}^m$.
\end{enumerate}
Then
\[
  \E\big[|E|\big] \;\ge\; \frac{2}{\ln 2}\, \varepsilon^2 N.
\]
\end{lemma}

\begin{proof}
Given $(\tilde{\rho}, E)$, define the reconstructed table $\widehat{T} \in \{0, 1\}^N$ by $\widehat{T}_{y \oplus c(\tilde{\rho})} = g(T, \tilde{\rho} \circ y)$ for all $y \in \{0, 1\}^m$. By Lemma~\ref{lem:pruning}, the expected normalized Hamming error between $\widehat{T}$ and $T$ is
\begin{align*}
  \delta \;:=\; \E\!\left[\frac{1}{N} \sum_{u \in \{0,1\}^m} \ind{\widehat{T}_u \neq T_u}\right]
  &\;=\; \Prob_{\tilde{\rho}, T, y}\!\Big[g(T, \tilde{\rho} \circ y) \neq T_{y \oplus c(\tilde{\rho})}\Big] \\
  &\;=\; \Prob_{T, x}\!\Big[g(T, x) \neq \Andreev_N(T, x)\Big]
  \;\le\; \frac{1}{2} - \varepsilon.
\end{align*}
Because $E$ takes values in a prefix-free set for every fixed $\tilde{\rho}_0$, Kraft's inequality and Gibbs' inequality give $\E[|E| \mid \tilde{\rho} = \tilde{\rho}_0] \ge H(E \mid \tilde{\rho} = \tilde{\rho}_0)$, so $\E[|E|] \ge H(E \mid \tilde{\rho}) \ge I(T; E \mid \tilde{\rho})$. Since $T \sim \Unif(\{0,1\}^N)$ is independent of $\tilde{\rho}$, we have $H(T \mid \tilde{\rho}) = N$. Moreover, since $\widehat{T}$ is a deterministic function of $(\tilde{\rho}, E)$,
\[
  H(T \mid \tilde{\rho}, E) \;=\; H(T \oplus \widehat{T} \mid \tilde{\rho}, E) \;\le\; H(T \oplus \widehat{T}) \;\le\; \sum_{u \in \{0,1\}^m} H(T_u \oplus \widehat{T}_u) \;=\; \sum_{u \in \{0,1\}^m} h_2(p_u),
\]
where $p_u = \Prob[T_u \neq \widehat{T}_u]$ and $h_2(q) = -q \log_2 q - (1-q) \log_2(1-q)$ is binary entropy. By concavity of $h_2$ and $\frac{1}{N} \sum_u p_u = \delta \le \frac{1}{2} - \varepsilon$,
\[
  \sum_{u \in \{0,1\}^m} h_2(p_u) \;\le\; N\, h_2(\delta) \;\le\; N\, h_2\!\left(\tfrac{1}{2} - \varepsilon\right).
\]
Using the standard Taylor bound $1 - h_2(\frac{1}{2} - \varepsilon) \ge \frac{2}{\ln 2} \varepsilon^2$ for $\varepsilon \in [0, \frac{1}{2}]$, we conclude
\[
  \E[|E|] \;\ge\; I(T; E \mid \tilde{\rho}) \;=\; N - H(T \mid \tilde{\rho}, E) \;\ge\; N\!\left(1 - h_2\!\left(\tfrac{1}{2} - \varepsilon\right)\right) \;\ge\; \frac{2}{\ln 2}\, \varepsilon^2 N. \qedhere
\]
\end{proof}

\subsection{Prefix-Free Encoding of the Final Layer}

A \emph{decision tree} $D$ over variables $y_1, \dots, y_m$ with output alphabet $\calH = \{0, 1\}^B$ is a rooted binary tree whose internal nodes are labeled by variables in $[m]$ (with no variable repeated along any root-to-leaf path), whose two outgoing edges correspond to values $0$ and $1$, and whose leaves are labeled by elements of $\calH$. Let $L(D) \ge 1$ denote the number of leaves of $D$; note that $D$ is constant if and only if $L(D) = 1$.

\begin{lemma}[Prefix-Free Layer-$L$ Encoding]
\label{lem:encoding}
Fix an $(L, k, H_{\mathrm{idx}}, B)$-Pairwise Sparse Transformer $g$, a table $T \in \{0, 1\}^N$, and a pruned Andreev restriction $\tilde{\rho}$. Suppose that for every position $t \in [n]$, the restricted Layer-$(L-1)$ state $y \mapsto h_t^{(L-1)}(T, \tilde{\rho} \circ y)$ is computed by a decision tree $D_t$ with $L_t = L(D_t)$ leaves. Let $A_{L-1}(\tilde{\rho}) = \{t \in [n] : L_t \ge 2\}$ be the set of non-constant Layer-$(L-1)$ tokens. Then there exists a prefix-free binary encoding $E(\tilde{\rho}, T)$ from which (together with $\tilde{\rho}$ and the fixed network $g$) one can deterministically reconstruct $y \mapsto g(T, \tilde{\rho} \circ y)$ for all $y \in \{0, 1\}^m$, of bit-length
\begin{align*}
  |E(\tilde{\rho}, T)| &\;\le\; \lceil \log_2(n+1) \rceil
  \;+\; \sum_{t \in A_{L-1}(\tilde{\rho})} L_t \big(B + \log_2 n + \log_2 m + 4\big) \\
  &\qquad {} +\; L_n \Big(B + \log_2 m + 3 + H_{\mathrm{idx}} k \big(B + \lceil \log_2(n+1) \rceil\big)\Big).
\end{align*}
\end{lemma}

\begin{proof}
First note that any binary decision tree $D$ over $m$ variables with $L(D)$ leaves has $L(D) - 1$ internal nodes, and admits a self-delimiting (prefix-free) preorder encoding: emit \texttt{1} followed by $\lceil \log_2 m \rceil$ bits specifying the queried variable for each internal node, and emit \texttt{0} followed by $B$ bits specifying the state in $\calH$ for each leaf. Its exact bit-length is
\[
  \big(L(D) - 1\big)\big(1 + \lceil \log_2 m \rceil\big) + L(D)(1 + B) \;\le\; L(D)\big(B + \log_2 m + 3\big).
\]

We construct $E(\tilde{\rho}, T)$ as the concatenation of three self-delimiting blocks:
\begin{enumerate}[leftmargin=2em,itemsep=2pt]
  \item \textbf{Count of non-constant tokens:} Encode $|A_{L-1}(\tilde{\rho})| \in \{0, \dots, n\}$ in $\lceil \log_2(n+1) \rceil$ bits.
  \item \textbf{Non-constant Layer-$(L-1)$ states:} For each $t \in A_{L-1}(\tilde{\rho})$, emit the index $t \in [n]$ in $\lceil \log_2 n \rceil$ bits followed by the self-delimiting preorder encoding of $D_t$. Since $L_t \ge 2$ for every $t \in A_{L-1}(\tilde{\rho})$, we have $\lceil \log_2 n \rceil \le L_t (\log_2 n + 1)$, so each record uses at most $L_t(B + \log_2 n + \log_2 m + 4)$ bits.
  \item \textbf{Final-layer query token $n$ and top-$k$ constant keys:} Emit the self-delimiting preorder encoding of $D_n$ (at most $L_n(B + \log_2 m + 3)$ bits). Then, for each leaf $\lambda$ of $D_n$ (in preorder, with state $u_\lambda \in \calH$) and each indexer head $r \in [H_{\mathrm{idx}}]$, consider all constant candidate keys $j \in \calC_n \setminus A_{L-1}(\tilde{\rho})$ (each having a constant state $h_j^{(L-1)} \in \calH$ independent of $y$). Let $K_{\lambda, r} \subseteq \calC_n \setminus A_{L-1}(\tilde{\rho})$ be the top $\min(k, |\calC_n \setminus A_{L-1}(\tilde{\rho})|)$ such indices ranked by $\big(\phi_{n, j}^{(L, r)}(u_\lambda, h_j^{(L-1)}), j\big)$ under $\succ_n$. Emit the pairs $(j, h_j^{(L-1)})$ for all $j \in K_{\lambda, r}$ (padding with dummy index $0$ if $|K_{\lambda, r}| < k$), using exactly $k(\lceil \log_2(n+1) \rceil + B)$ bits per $(\lambda, r)$.
\end{enumerate}
Summing the lengths yields the stated bound, and every block is self-delimiting, so $E(\tilde{\rho}, T)$ is prefix-free.

\emph{Correctness of decoding.} Fix any $y \in \{0, 1\}^m$. Using block~3, the decoder evaluates $D_n(y)$ to reach a leaf $\lambda$ and obtain $u_\lambda = h_n^{(L-1)}(T, \tilde{\rho} \circ y)$. Using block~2, it evaluates $D_t(y)$ for every $t \in A_{L-1}(\tilde{\rho})$ to obtain $h_t^{(L-1)}(T, \tilde{\rho} \circ y)$. Now consider any indexer head $r \in [H_{\mathrm{idx}}]$. Every constant key $j \in \calC_n \setminus A_{L-1}(\tilde{\rho})$ outside $K_{\lambda, r}$ is strictly beaten under $\succ_n$ by all $k$ keys in $K_{\lambda, r}$ whenever $h_n^{(L-1)} = u_\lambda$, and therefore can \emph{never} enter $S_n^{(L, r)}(T, \tilde{\rho} \circ y)$. Consequently,
\[
  S_n^{(L, r)}(T, \tilde{\rho} \circ y) \;=\; \TopK_{j \in K_{\lambda, r} \cup (\calC_n \cap A_{L-1}(\tilde{\rho}))}\!\left(\phi_{n, j}^{(L, r)}\!\big(u_\lambda, h_j^{(L-1)}(T, \tilde{\rho} \circ y)\big),\, k;\, \prec_n\right).
\]
Since the decoder knows $h_j^{(L-1)}(T, \tilde{\rho} \circ y)$ for every $j \in K_{\lambda, r} \cup A_{L-1}(\tilde{\rho})$, it computes the exact set $S_n^{(L)}(T, \tilde{\rho} \circ y)$ and all selected states, and applies the fixed network functions $\psi_n^{(L)}$ and $\Out$ to recover $g(T, \tilde{\rho} \circ y)$.
\end{proof}

\section{Proof of Theorem~\ref{thm:main}: Top-\texorpdfstring{$k$}{k} Switching and Multi-Layer Collapse}
\label{sec:proof-main}

We decompose the random restriction $\rho \sim \calR_p$ on $V_x$ into $L$ independent stages
\[
  \rho \;=\; \rho_1 \circ \rho_2 \circ \dots \circ \rho_{L-1} \circ \rho_{\mathrm{final}},
\]
where $\rho_\ell \sim \calR_{p_\ell}$ collapses Layer~$\ell$ to shallow decision trees via our Top-$k$ Switching Lemma ($\ell = 1, \dots, L-1$), and $\rho_{\mathrm{final}} \sim \calR_{p_{\mathrm{final}}}$ shrinks almost all Layer-$(L-1)$ decision trees to constants.

\subsection{A Switching Lemma for Finding the First \texorpdfstring{$k$}{k} Satisfied Conjunctions}

Let $\mathcal{T} = (T_1, T_2, \dots, T_M)$ be an ordered list of satisfiable conjunctions over a set of Boolean variables $V$, each of width at most $W$ (a conjunction of at most $W$ literals over distinct variables; the empty conjunction is identically $1$). Given a restriction $\rho : V \to \{0, 1, *\}$, we define the \emph{canonical Top-$k$ decision tree} $\CDT_k(\mathcal{T}; \rho)$, which determines the ordered tuple of the first $k$ indices $i \in [M]$ for which $T_i = 1$ (or all such indices if fewer than $k$ evaluate to $1$):
\begin{quote}
  Initialize $\mathrm{hits} \leftarrow 0$, $\mathrm{idx} \leftarrow 0$, and partial assignment $\sigma \leftarrow \emptyset$. While $\mathrm{hits} < k$:
  \begin{enumerate}[leftmargin=2em,itemsep=1pt]
    \item Let $i$ be the smallest index strictly greater than $\mathrm{idx}$ such that $T_i$ is \emph{not falsified} by $\rho \circ \sigma$. If no such $i$ exists, halt.
    \item If $T_i$ is already \emph{satisfied} by $\rho \circ \sigma$, record $i$, set $\mathrm{hits} \leftarrow \mathrm{hits} + 1$, $\mathrm{idx} \leftarrow i$, and continue ($0$ queries).
    \item Otherwise, let $U_i$ be the nonempty set of variables of $T_i$ that remain $*$ under $\rho \circ \sigma$ ($1 \le |U_i| \le W$). Query all variables in $U_i$ in canonical order, extending $\sigma$ with the answers. If the answers satisfy $T_i$, record $i$ and set $\mathrm{hits} \leftarrow \mathrm{hits} + 1$. Set $\mathrm{idx} \leftarrow i$.
  \end{enumerate}
\end{quote}

\begin{lemma}[Top-$k$ Switching Lemma]
\label{lem:topk-switching}
Let $\mathcal{T} = (T_1, \dots, T_M)$ be any ordered list of satisfiable conjunctions of width at most $W$. Then for every $p \in (0, 1)$ and every integer $r \ge 1$,
\[
  \Prob_{\rho \sim \calR_p}\!\Big[\operatorname{depth}\!\big(\CDT_k(\mathcal{T}; \rho)\big) \ge r\Big]
  \;\le\; 2^k \left(\frac{16 p W}{1 - p}\right)^r.
\]
\end{lemma}

\begin{proof}
We generalize Razborov's encoding proof of H\aa stad's Switching Lemma~\cite{razborov1995bounded,beame1994switching} from $k = 1$ to arbitrary $k \ge 1$. If $W = 0$, every conjunction is empty and $\CDT_k$ makes $0$ queries, so the claim is trivial; assume $W \ge 1$. Let $\mathcal{B} = \{\rho : \operatorname{depth}(\CDT_k(\mathcal{T}; \rho)) \ge r\}$. For each $\rho \in \mathcal{B}$, fix the lexicographically first root-to-leaf path $\pi$ in $\CDT_k(\mathcal{T}; \rho)$ of length at least $r$, and truncate $\pi$ at \emph{exactly} $r$ queried variables.

Let $T_{i_1}, T_{i_2}, \dots, T_{i_b}$ ($1 \le b \le r$) be the conjunctions queried along this truncated path, where in block $m \in [b]$ the tree queries a nonempty set $V_m \subseteq \mathrm{vars}(T_{i_m})$ of $a_m = |V_m| \ge 1$ variables, with $\sum_{m=1}^b a_m = r$. Note that $V_1, \dots, V_b$ are pairwise disjoint subsets of $\{v \in V : \rho(v) = *\}$. For each $m \in [b]$:
\begin{itemize}[leftmargin=1.5em,itemsep=1pt]
  \item let $\beta_m \in \{0, 1\}^{V_m}$ be the bit assignment followed by $\pi$;
  \item let $\alpha_m \in \{0, 1\}^{V_m}$ be the unique assignment to $V_m$ consistent with $T_{i_m} = 1$; and
  \item let $h_m \ge 0$ be the number of indices $i' \in (i_{m-1}, i_m)$ (with $i_0 = 0$) such that $T_{i'}$ was already satisfied by $\rho \circ \beta_1 \circ \dots \circ \beta_{m-1}$.
\end{itemize}
Because every already-satisfied term increments $\mathrm{hits}$ and $\mathrm{hits} \le k - 1$ upon entering the final block $b$, we have $\sum_{m=1}^b h_m \le k - 1$.

Define the modified restriction
\[
  \rho' \;=\; \rho \circ \alpha_1 \circ \alpha_2 \circ \dots \circ \alpha_b,
\]
which fixes exactly $r$ additional variables that were $*$ in $\rho$. Define the advice string $\mathrm{adv}(\rho)$ consisting of:
\begin{enumerate}[label=(\roman*),leftmargin=2em,itemsep=1pt]
  \item the tuple $(h_1, \dots, h_b, 0, \dots, 0) \in \mathbb{Z}_{\ge 0}^r$ (padded to length $r$) with $\sum_{m=1}^r h_m \le k - 1$; and
  \item for each of the $r$ queried variables (ordered across blocks $1, \dots, b$): its position in the canonical literal list of $T_{i_m}$ (at most $W$ choices), a $1$-bit flag indicating whether it ends block $m$ ($2$ choices), and its value under $\beta_m$ ($2$ choices).
\end{enumerate}

\emph{Injectivity of $\rho \mapsto (\rho', \mathrm{adv}(\rho))$.} We show that $\rho$ is uniquely recoverable from $(\rho', \mathrm{adv}(\rho))$. Set $i_0 = 0$ and $\rho'_0 = \rho'$. Inductively for $m = 1, \dots, b$, suppose the decoder knows $i_{m-1}$ and holds
\[
  \rho'_{m-1} \;=\; \rho \circ \beta_1 \circ \dots \circ \beta_{m-1} \circ \alpha_m \circ \dots \circ \alpha_b.
\]
Consider all indices $i' \in (i_{m-1}, i_m)$. By definition of $\CDT_k$, under $\rho \circ \beta_1 \circ \dots \circ \beta_{m-1}$ every such $T_{i'}$ was either falsified (in which case it remains falsified in $\rho'_{m-1}$, since $\alpha_m \circ \dots \circ \alpha_b$ only assigns variables that were $*$ under $\rho \circ \beta_1 \circ \dots \circ \beta_{m-1}$) or satisfied (in which case it remains satisfied—hence not falsified—in $\rho'_{m-1}$; there are \emph{exactly} $h_m$ such indices). Meanwhile, $T_{i_m}$ itself was not falsified under $\rho \circ \beta_1 \circ \dots \circ \beta_{m-1}$:
\begin{itemize}[leftmargin=1.5em,itemsep=1pt]
  \item if $m < b$, then $V_m$ contains all remaining alive variables of $T_{i_m}$, so $\alpha_m$ satisfies $T_{i_m}$ completely, and hence $\alpha_{m+1}, \dots, \alpha_b$ cannot touch $\mathrm{vars}(T_{i_m})$; thus $T_{i_m}$ is satisfied (not falsified) in $\rho'_{m-1}$;
  \item if $m = b$, then $\rho'_{b-1} = \rho \circ \beta_1 \circ \dots \circ \beta_{b-1} \circ \alpha_b$, and $\alpha_b$ agrees with $T_{i_b}$ on $V_b$ while leaving any unqueried alive variables of $T_{i_b}$ as $*$; thus $T_{i_b}$ is not falsified in $\rho'_{b-1}$.
\end{itemize}
Therefore, $T_{i_m}$ is uniquely identified as the $(h_m + 1)$-st conjunction after $i_{m-1}$ that is not falsified by $\rho'_{m-1}$. Reading the block-$m$ entries of item~(ii) identifies $V_m$ and $\beta_m$, allowing the decoder to overwrite $\alpha_m$ with $\beta_m$ to form $\rho'_m$ and continue until all $r$ queried variables have been processed. Finally, unsetting all $r$ variables in $V_1 \cup \dots \cup V_b$ back to $*$ recovers $\rho$ uniquely.

\emph{Probability accounting.} For every $\rho \in \mathcal{B}$, since $\rho'$ fixes $r$ coordinates that were $*$ in $\rho$,
\[
  \Prob[\rho] \;=\; \left(\frac{p}{(1-p)/2}\right)^r \Prob[\rho'] \;=\; \left(\frac{2p}{1-p}\right)^r \Prob[\rho'].
\]
The number of choices for item~(i) is $\binom{r + k - 1}{k - 1} \le 2^{r + k - 1} \le 2^{r + k}$, and the number of choices for item~(ii) is at most $(4 W)^r$, so the total number of advice strings is at most $2^k (8 W)^r$. By injectivity,
\[
  \Prob_{\rho \sim \calR_p}[\mathcal{B}] \;=\; \sum_{\rho \in \mathcal{B}} \Prob[\rho]
  \;=\; \left(\frac{2p}{1-p}\right)^r \sum_{\rho \in \mathcal{B}} \Prob[\rho']
  \;\le\; \left(\frac{2p}{1-p}\right)^r \cdot 2^k (8 W)^r \sum_{\rho'} \Prob[\rho']
  \;=\; 2^k \left(\frac{16 p W}{1 - p}\right)^r. \qedhere
\]
\end{proof}

\subsection{Multi-Layer Decision-Tree Collapse and Leaf Shrinkage}

\begin{lemma}[Multi-Layer Collapse]
\label{lem:multi-collapse}
Fix any $(L, k, H_{\mathrm{idx}}, B)$-Pairwise Sparse Transformer with $L \ge 2$ and any table $T \in \{0, 1\}^N$. Set
\[
  w \;=\; H_{\mathrm{idx}}\Big(k + \big\lceil 10 \log_2(n H_{\mathrm{idx}}) \big\rceil\Big),
  \qquad
  p_1 \;=\; \frac{1}{160},
  \qquad
  p_2 \;=\; \dots \;=\; p_{L-1} \;=\; \frac{1}{160 w},
\]
and let $\rho_{\mathrm{switch}} = \rho_1 \circ \dots \circ \rho_{L-1} \sim \calR_{p_{\mathrm{switch}}}$ with $p_{\mathrm{switch}} = \frac{1}{160^{L-1} w^{L-2}}$. Then with probability at least $1 - (L - 1) n^{-19}$ (denote this event by $\calG_{L-1}$), every Layer-$(L-1)$ state $h_t^{(L-1)}(T, \rho_{\mathrm{switch}} \circ \cdot)$ ($t \in [n]$) is computed by a decision tree of depth at most $w$ over the surviving variables of $V_x$.
\end{lemma}

\begin{proof}
We induct on $\ell = 0, 1, \dots, L - 1$. For $\ell = 0$, each $h_t^{(0)}$ is a decision tree of depth $w_0 \le 1$ over $V_x$ ($0$ for $t \in [N]$, $1$ for $t \in V_x$).

For the step $\ell - 1 \to \ell$ ($\ell \in \{1, \dots, L-1\}$), condition on the inductive event $\calG_{\ell-1}$ that every $h_j^{(\ell-1)}|_{\rho_1 \cdots \rho_{\ell-1}}$ ($j \in [n]$) is computed by a decision tree $D_j$ of depth at most $w_{\ell-1}$ (where $w_0 = 1$ and $w_{\ell-1} = w$ for $\ell \ge 2$). Fix any position $t \in [n]$ and any indexer head $r \in [H_{\mathrm{idx}}]$. For each candidate key $j \in \calC_t$, form the combined tree $D_{t, j}$ of depth at most $W_\ell = 2 w_{\ell-1}$ by attaching a copy of $D_j$ at every leaf of $D_t$ (pruning redundant queries). Each leaf $\lambda$ of $D_{t, j}$ corresponds to a satisfiable conjunction $T_{j, \lambda}$ of width at most $W_\ell$ over the surviving $V_x$ variables and uniquely determines $\big(h_t^{(\ell-1)}, h_j^{(\ell-1)}\big)$ as well as the score $I_{t, j}^{(\ell, r)} = \phi_{t, j}^{(\ell, r)}(h_t^{(\ell-1)}, h_j^{(\ell-1)})$.

Collect all leaves $\{(j, \lambda) : j \in \calC_t,\; \lambda \in \mathrm{leaves}(D_{t, j})\}$ into a single master list $\mathcal{T}_{t, r} = (T_1, \dots, T_M)$ sorted in decreasing order of $\big(I_{t, j}^{(\ell, r)}, j\big)$ under $\succ_t$ (breaking ties among distinct leaves of the same $j$ arbitrarily). On every completion of the surviving variables, for each $j \in \calC_t$ \emph{exactly one} leaf of $D_{t, j}$ evaluates to $1$ (and all true leaves agree on the same value of $h_t^{(\ell-1)}$). Consequently, the first $\min(k, |\calC_t|)$ conjunctions in $\mathcal{T}_{t, r}$ that evaluate to $1$ determine the exact Top-$k$ set $S_t^{(\ell, r)}$, the query state $h_t^{(\ell-1)}$, and all selected states $\{(j, h_j^{(\ell-1)}) : j \in S_t^{(\ell, r)}\}$.

Apply Lemma~\ref{lem:topk-switching} to $\mathcal{T}_{t, r}$ under $\rho_\ell \sim \calR_{p_\ell}$ with $r_0 = k + \lceil 10 \log_2(n H_{\mathrm{idx}}) \rceil$. Since $p_1 W_1 = \frac{2}{160}$ and $p_\ell W_\ell = \frac{2 w}{160 w} = \frac{2}{160}$ for $\ell \ge 2$, we have $\frac{16 p_\ell W_\ell}{1 - p_\ell} \le \frac{32}{159} < \frac{1}{4}$ in every layer, and hence
\[
  \Prob_{\rho_\ell}\!\Big[\operatorname{depth}\!\big(\CDT_k(\mathcal{T}_{t, r}; \rho_\ell)\big) \ge r_0\Big]
  \;\le\; 2^k \cdot 4^{-r_0}
  \;\le\; (n H_{\mathrm{idx}})^{-20}.
\]
By a union bound over all $n H_{\mathrm{idx}}$ pairs $(t, r)$, with probability at least $1 - (n H_{\mathrm{idx}})^{-19} \ge 1 - n^{-19}$, every $\CDT_k(\mathcal{T}_{t, r}; \rho_\ell)$ has depth at most $r_0$. Concatenating the $H_{\mathrm{idx}}$ canonical trees $\{\CDT_k(\mathcal{T}_{t, r}; \rho_\ell)\}_{r=1}^{H_{\mathrm{idx}}}$ (and pruning any repeated variable along a root-to-leaf path) yields a decision tree of depth at most $H_{\mathrm{idx}} r_0 = w$ computing $h_t^{(\ell)}|_{\rho_1 \cdots \rho_\ell}$ for every $t \in [n]$. Summing the failure probabilities over $\ell = 1, \dots, L - 1$ completes the induction.
\end{proof}

\begin{lemma}[Decision-Tree Leaf Shrinkage under Final Restriction]
\label{lem:leaf-shrinkage}
Let $D$ be any decision tree of depth at most $w$ over variables $V$, and let $\rho \sim \calR_q$ on $V$. Let $L(D|_\rho)$ denote the number of leaves of the simplified decision tree $D|_\rho$ obtained by replacing each fixed internal node by its selected child. Then
\[
  \E_{\rho \sim \calR_q}\!\big[L(D|_\rho)\big] \;\le\; (1 + q)^w.
\]
Moreover, if $w q \le \frac{1}{2}$, then
\[
  \E_{\rho \sim \calR_q}\!\Big[\ind{L(D|_\rho) \ge 2}\, L(D|_\rho)\Big] \;\le\; 4 w q.
\]
\end{lemma}

\begin{proof}
We prove $\E[L(D|_\rho)] \le (1 + q)^w$ by induction on $w$. If $w = 0$, $D$ is a single leaf and $L(D|_\rho) = 1 = (1 + q)^0$. For $w \ge 1$, let $v \in V$ be the variable queried at the root of $D$, and let $D_0, D_1$ be its two subtrees of depth at most $w - 1$. Because $v$ does not occur in $D_0$ or $D_1$, $\rho(v)$ is independent of the restriction of $\rho$ to $D_0$ and $D_1$. Conditioning on $\rho(v)$:
\begin{itemize}[leftmargin=1.5em,itemsep=1pt]
  \item if $\rho(v) = *$ (prob $q$), then $L(D|_\rho) = L(D_0|_\rho) + L(D_1|_\rho)$;
  \item if $\rho(v) = b \in \{0, 1\}$ (prob $\frac{1-q}{2}$ each), then $L(D|_\rho) = L(D_b|_\rho)$.
\end{itemize}
Taking expectations and applying the induction hypothesis yields
\begin{align*}
  \E\big[L(D|_\rho)\big]
  &\;=\; \left(q + \frac{1 - q}{2}\right)\Big(\E[L(D_0|_\rho)] + \E[L(D_1|_\rho)]\Big) \\
  &\;=\; \frac{1 + q}{2}\Big(\E[L(D_0|_\rho)] + \E[L(D_1|_\rho)]\Big)
  \;\le\; (1 + q)^w.
\end{align*}
For the second claim, note that $\ind{\ell \ge 2}\,\ell \le 2(\ell - 1)$ for every integer $\ell \ge 1$. When $w q \le \frac{1}{2}$, we have $(1 + q)^w \le e^{wq}$ and $e^u - 1 \le 2u$ on $[0, \frac{1}{2}]$, hence
\[
  \E\Big[\ind{L(D|_\rho) \ge 2}\, L(D|_\rho)\Big]
  \;\le\; 2\,\E\big[L(D|_\rho) - 1\big]
  \;\le\; 2\big((1 + q)^w - 1\big)
  \;\le\; 2\big(e^{wq} - 1\big)
  \;\le\; 4 w q. \qedhere
\]
\end{proof}

\subsection{Completing the Proof of Theorem~\ref{thm:main}}

\begin{proof}[Proof of Theorem~\ref{thm:main}]
As noted at the beginning of Section~\ref{sec:framework}, for $N < 32$ the theorem holds trivially by adjusting the hidden constant in $\Omega(\cdot)$, so assume $N = 2^m \ge 32$ ($m \ge 5$). Let $w = H_{\mathrm{idx}}\big(k + \lceil 10 \log_2(n H_{\mathrm{idx}}) \rceil\big)$ and $p = \frac{2\ln(2m)}{s} \in (0, 1)$ as in~\eqref{eq:p-def}. Note that if $L - 1 \ge \log_2 N$, then $(160 w)^{L-1} \ge 160^{\log_2 N} > N$, so the theorem holds trivially. Assume therefore that $L \le \log_2 N$.

Set $p_{\mathrm{switch}} = \frac{1}{160^{L-1} w^{L-2}}$ as in Lemma~\ref{lem:multi-collapse}, and define
\[
  p_{\mathrm{final}} \;=\; \frac{p}{p_{\mathrm{switch}}} \;=\; 160^{L-1} w^{L-2} p.
\]
If $w p_{\mathrm{final}} > \frac{1}{2}$, then
\[
  (160 w)^{L-1} \;>\; \frac{1}{2p} \;=\; \frac{s}{4 \ln(2m)} \;=\; \Omega\!\left(\frac{N}{\log N \log\log N}\right),
\]
which already establishes Theorem~\ref{thm:main} (since $B + \log_2 N \ge 1$ and $\varepsilon \le \frac{1}{2}$).

Assume therefore that $w p_{\mathrm{final}} \le \frac{1}{2}$ (in particular $p_{\mathrm{final}} \le \frac{1}{2} < 1$). Then composing $\rho_{\mathrm{switch}} \sim \calR_{p_{\mathrm{switch}}}$ with an independent $\rho_{\mathrm{final}} \sim \calR_{p_{\mathrm{final}}}$ produces $\rho = \rho_{\mathrm{switch}} \circ \rho_{\mathrm{final}} \sim \calR_p$.

Fix $T \in \{0, 1\}^N$ and sample $(\rho_{\mathrm{switch}}, \rho_{\mathrm{final}})$ conditioned on $\calE_{\mathrm{alive}}$ (which has probability $\ge \frac{1}{2}$ by~\eqref{eq:e-alive}), then prune $\rho$ to $\tilde{\rho}$. Construct a (randomized, via $\rho_{\mathrm{switch}}$) prefix-free encoding $E^*$ of $y \mapsto g(T, \tilde{\rho} \circ y)$ using one initial flag bit:
\begin{itemize}[leftmargin=1.5em,itemsep=2pt]
  \item If $\calG_{L-1}$ fails (flag \texttt{1}): emit the entire table $T$ verbatim ($N$ bits). Since $\Prob[\neg \calG_{L-1} \mid \calE_{\mathrm{alive}}] \le 2 \Prob[\neg \calG_{L-1}] \le 2(L - 1) n^{-19} \le 2 N^{-18}$ (as $L \le \log_2 N \le n$), the expected contribution of these $N$ table bits to $|E^*|$ is at most $2 N^{-17} \le 1$ bit.
  \item If $\calG_{L-1}$ holds (flag \texttt{0}): every Layer-$(L-1)$ state $h_t^{(L-1)}|_{\rho_{\mathrm{switch}}}$ is a decision tree $D_t^{(0)}$ of depth at most $w$. Applying $\rho_{\mathrm{final}}$ and then pruning $\rho \to \tilde{\rho}$ produces simplified decision trees $D_t = D_t^{(0)}|_{\tilde{\rho}}$ whose leaf counts satisfy $L_t = L(D_t) \le L(D_t^{(0)}|_{\rho_{\mathrm{final}}})$ point-wise (pruning only replaces internal nodes by children and never increases leaf counts). Emit the prefix-free encoding $E(\tilde{\rho}, T)$ from Lemma~\ref{lem:encoding}.
\end{itemize}
Observe that for every fixed $\tilde{\rho}_0$, every possible output of $E^*$ is either \texttt{1} followed by $N$ bits or \texttt{0} followed by a self-delimiting codeword from Lemma~\ref{lem:encoding}, so the support of $E^*$ given $\tilde{\rho} = \tilde{\rho}_0$ is prefix-free; moreover, decoding $(\tilde{\rho}, E^*)$ requires only $\tilde{\rho}$, $E^*$, and the fixed network $g$ (never $\rho_{\mathrm{switch}}$).

Because $\rho_{\mathrm{final}}$ is independent of $\rho_{\mathrm{switch}}$ (and hence of $\calG_{L-1}$), Lemma~\ref{lem:leaf-shrinkage} applies directly conditioned on $\calG_{L-1}$: for every $t \in [n]$,
\begin{align*}
  \E_{\rho_{\mathrm{final}}}\!\Big[\ind{L(D_t^{(0)}|_{\rho_{\mathrm{final}}}) \ge 2}\, L(D_t^{(0)}|_{\rho_{\mathrm{final}}}) \;\Big|\; \calG_{L-1}\Big] &\;\le\; 4 w p_{\mathrm{final}}, \\
  \E_{\rho_{\mathrm{final}}}\!\big[L(D_n^{(0)}|_{\rho_{\mathrm{final}}}) \;\big|\; \calG_{L-1}\big] &\;\le\; e^{1/2} \;<\; 2.
\end{align*}
Summing Lemma~\ref{lem:encoding} over all $t \in [n]$ and dividing by $\Prob[\calE_{\mathrm{alive}}] \ge \frac{1}{2}$ gives, for every $T \in \{0,1\}^N$,
\begin{align*}
  \E\big[|E^*| \mid T\big]
  &\;\le\; 2 + 2\Big[\lceil \log_2(n+1) \rceil
    + (4 n w p_{\mathrm{final}})(B + \log_2 n + \log_2 m + 4) \\
  &\qquad\qquad\quad {} + 2\Big(B + \log_2 m + 3 + H_{\mathrm{idx}} k \big(B + \lceil \log_2(n+1) \rceil\big)\Big)\Big].
\end{align*}
Substituting $n w p_{\mathrm{final}} = 160^{L-1} w^{L-1} (n p) = O\!\left(160^{L-1} w^{L-1} \log N \log\log N\right)$ (using $n \le 2N$ and $n p \le 2N \cdot \frac{2\ln(2m)}{\lfloor N/m \rfloor} = O(\log N \log\log N)$) and noting $H_{\mathrm{idx}} k \le w \le 160^{L-1} w^{L-1}$ yields
\[
  \E\big[|E^*|\big]
  \;=\; O\!\Big(160^{L-1} w^{L-1} (B + \log N) \log N \log\log N\Big).
\]
Combining this upper bound with Lemma~\ref{lem:fano} gives
\[
  \frac{2}{\ln 2}\, \varepsilon^2 N \;\le\; O\!\Big(160^{L-1} w^{L-1} (B + \log N) \log N \log\log N\Big).
\]
Since $n \le 2 N$ and $w = H_{\mathrm{idx}}\big(k + \lceil 10 \log_2(n H_{\mathrm{idx}}) \rceil\big) = O\!\big(H_{\mathrm{idx}}(k + \log_2(N H_{\mathrm{idx}}))\big)$, choosing an appropriate absolute constant $C \ge 1$ finishes the proof.
\end{proof}

\section{Discussion and Open Problems}
\label{sec:discussion}

Our results show that the trade-off between full and sparse attention is not merely an engineering trade-off of FLOPs versus benchmark noise—it is a sharp, unconditional average-case computational separation even well inside the supercritical connectivity regime $k^L > n$. We conclude with three natural open questions suggested by our work:

\begin{enumerate}[leftmargin=2em,itemsep=3pt]
  \item \textbf{Bypassing the $\mathsf{TC}^0$ barrier via realistic attention constraints.}
  As noted in Remark~\ref{rem:tc0-barrier}, Definition~\ref{def:sparse-transformer} allows $\psi_t^{(\ell)}$ to implement arbitrary signed MLPs, which makes our model powerful enough to simulate general $\mathsf{TC}^0$ circuits. Real Transformers mix selected value vectors via \emph{non-negative} softmax attention weights followed by low-depth, bounded-norm MLPs. Can one exploit non-negativity or low-rank value projections to prove stronger average-case lower bounds for sparse attention at larger depths ($L \ge 4$ with $k^2 \gg N$)?

  \item \textbf{Hybrid architectures and cross-layer index reuse.}
  Many recent open-weight models interleave a minority of full-attention layers among sparse or linear layers, or reuse Top-$k$ indices across blocks of $r$ consecutive layers (as in DeepSeek-V4.1's CSA2, GLM-5.2's IndexShare, and Hunyuan's IndexCache). Formalizing how cross-layer index reuse shrinks effective depth—and characterizing the exact number of full-attention layers needed to close the representational gap on multi-hop variants of Andreev's function—is an exciting direction for both theory and architecture design.

  \item \textbf{Optimization bottlenecks on top of representational limits.}
  Theorem~\ref{thm:main} is purely \emph{representational}: it holds even for an omniscient optimizer with unlimited training data. In practice, training non-differentiable Top-$k$ indexers via surrogate KL-alignment losses introduces severe gradient approximation errors and exploration failures. Quantifying how optimization difficulty compounds the representational gap we prove here may explain an even larger share of the empirical divergence between open-source sparse-attention LLMs and closed-source frontier models.
\end{enumerate}

\appendix
\section{Proofs of Upper Bounds (Propositions~\ref{prop:full-upper} and~\ref{prop:sparse-upper})}
\label{app:upper-bounds}

\begin{proof}[Proof of Proposition~\ref{prop:full-upper}]
Let $n = N + m s$ with $m = \log_2 N$.

\emph{Part (a): $2$-layer causal Full Attention with width-$O(N)$ ReLU MLP.}
Use $H = 1$ head, embedding dimension $d = m + 3 = O(\log N)$, and $O(\log N)$-bit precision.
\begin{itemize}[leftmargin=1.5em,itemsep=2pt]
  \item \emph{Layer 1:} Set all queries and keys identically to $0$, so causal softmax attention at token $n$ assigns exact uniform weight $\frac{1}{n}$ to every token $j \in [n]$. Set value vectors $v_j^{(1)} = 0 \in \R^m$ for $j \in [N]$, and $v_j^{(1)} = x_j e_r \in \R^m$ for $j \in G_r$ ($r \in [m]$), where $e_r$ is the $r$-th standard basis vector. The Layer-$1$ attention output at position $n$ is
  \[
    a_n^{(1)} \;=\; \frac{1}{n}\!\left(\sum_{j \in G_1} x_j,\; \dots,\; \sum_{j \in G_m} x_j\right).
  \]
  Rescaling by $n$ recovers the exact integer block sums $c_r = \sum_{j \in G_r} x_j \in \{0, 1, \dots, s\}$. For each $r \in [m]$, a $2$-layer ReLU subnetwork of width $s + 1$ computes $S_r(x) = c_r \bmod 2$ via the exact integer identity
  \[
    c \bmod 2 \;=\; \ReLU(c) \;-\; 2 \sum_{a = 1}^{s} (-1)^{a - 1} \ReLU(c - a) \qquad \forall\, c \in \{0, 1, \dots, s\},
  \]
  requiring total MLP width $m(s + 1) \le N + m = O(N)$. Thus token $n$ stores $S(x) \in \{0, 1\}^m$ after Layer~1.

  \item \emph{Layer 2:} Set $\beta = \ln(4n) = O(\log N)$. At token $n$, form the query $q_n^{(2)} = \beta\big(2 S(x) - \mathbf{1}_m,\, 1\big) \in \R^{m+1}$. For each table token $j \in [N]$ with binary index $u = \bin(j) \in \{0, 1\}^m$, set the positional key $k_j^{(2)} = (2 u - \mathbf{1}_m,\, 1) \in \{-1, +1\}^{m+1}$ and value $v_j^{(2)} = T_j \in \{0, 1\}$. For each address token $j \in \{N+1, \dots, n\}$, set $k_j^{(2)} = (0, \dots, 0, -m) \in \R^{m+1}$ and $v_j^{(2)} = 0$.

  Let $j^* \in [N]$ be the unique table token with $\bin(j^*) = S(x)$. Then $\langle q_n^{(2)}, k_{j^*}^{(2)} \rangle = \beta(m + 1)$, whereas for every $j \in [N] \setminus \{j^*\}$ we have $\langle q_n^{(2)}, k_j^{(2)} \rangle \le \beta(m - 1)$, and for every $j > N$ we have $\langle q_n^{(2)}, k_j^{(2)} \rangle = -\beta m \le \beta(m - 1)$. Hence the softmax weight on $j^*$ satisfies
  \[
    \alpha_{n, j^*}^{(2)} \;\ge\; \frac{1}{1 + (n - 1) e^{-2\beta}} \;=\; \frac{1}{1 + \frac{n-1}{16 n^2}} \;>\; 1 - \frac{1}{16 n}.
  \]
  Since all values lie in $\{0, 1\}$, the attention output $a_n^{(2)}$ satisfies $|a_n^{(2)} - T_{S(x)}| < \frac{1}{16 n} < \frac{1}{2}$, and thresholding at $\frac{1}{2}$ outputs $T_{S(x)}$ exactly.
\end{itemize}

\emph{Part (b): $3$-layer causal Full Attention with width-$O(\log N)$ ReLU MLP.}
Keep Layer~1's uniform attention to compute $(c_1, \dots, c_m)$ at token $n$ with a linear projection (no large MLP). In Layer~2, use $m = \log_2 N$ heads. For head $r \in [m]$, set $\beta_2 = \ln(4n)$ and query $q_{n, r}^{(2)} = (2\beta_2 c_r,\, -\beta_2,\, \beta_2) \in \R^3$; for positions $j \in \{1, \dots, s+1\}$ (setting $t = j - 1 \in \{0, \dots, s\}$, noting $s + 1 \le N$), set positional key $k_{j, r}^{(2)} = (t,\, t^2,\, 0)$ and value $v_{j, r}^{(2)} = t \bmod 2$; for $j > s + 1$, set $k_{j, r}^{(2)} = (0, 0, -1)$ and $v_{j, r}^{(2)} = 0$. For $j \in \{1, \dots, s+1\}$,
\[
  \langle q_{n, r}^{(2)}, k_{j, r}^{(2)} \rangle \;=\; \beta_2(2 c_r t - t^2) \;=\; \beta_2 c_r^2 - \beta_2(c_r - t)^2,
\]
which uniquely attains $\beta_2 c_r^2$ at $t = c_r$ and is at most $\beta_2 c_r^2 - \beta_2$ for all other $j \in [n]$ (including $j > s+1$, which has logit $-\beta_2 \le \beta_2 c_r^2 - \beta_2$). Thus head $r$'s softmax weight on $j = c_r + 1$ exceeds $\frac{1}{1 + (n-1)e^{-\beta_2}} > \frac{3}{4}$, so thresholding each of the $m$ coordinates at $\frac{1}{2}$ (width $O(\log N)$ MLP) recovers $S(x)$ exactly. Layer~3 then performs the exact same table lookup as Layer~2 in part~(a).
\end{proof}

\begin{proof}[Proof of Proposition~\ref{prop:sparse-upper}]
\emph{Part (a): $k^{L-1} \ge N$ and $B \ge \log_2 N$.}
Since $|V_x| = m s \le N \le k^{L-1}$, organize $V_x$ as the leaves of a fixed positional $k$-ary reduction tree of depth $L - 1$, placing each internal node at the rightmost token of its subtree (so every child precedes its parent, respecting causal masking $\calC_t = [t]$, and the root lies at position $n$). Using purely positional pairwise scores in Layers $1, \dots, L - 1$, each internal node at depth $\ell$ attends to its at most $k$ children from layer $\ell - 1$ and computes the coordinatewise XOR of their $m$-bit vectors of partial parities across all $m$ blocks $G_1, \dots, G_m$ (which fits in $m = \log_2 N \le B$ bits). After Layer~$L - 1$, token $n$ holds the exact $m$-bit address $S(x)$. In Layer~$L$, token $n$ uses a pairwise indexer matching $S(x)$ against $\bin(j)$ (with $k = 1$) to select $j^* = S(x) \in [N]$ and output $T_{S(x)}$.

\emph{Part (b): $L \ge 2$, $k \ge 4$, $k^L \ge 4 N$, $k B \ge 4 N$, and $B \ge \log_2 N$.}
Set $B' = \min(B, k^{L-1})$, so $k B' = \min(k B, k^L) \ge 4 N$. Partition $[N]$ into $M_T = \lceil N / B' \rceil$ contiguous blocks of size at most $B' \le k^{L-1}$, and partition $V_x$ into $M_x = \lceil m s / k^{L-1} \rceil$ contiguous blocks of size at most $k^{L-1}$. Since $N / B' \le k / 4$, $m s / k^{L-1} \le N / k^{L-1} \le k / 4$, and $k \ge 4$, we have
\[
  M_T + M_x \;\le\; \left(\frac{k}{4} + 1\right) + \left(\frac{k}{4} + 1\right) \;=\; \frac{k}{2} + 2 \;\le\; k.
\]
Inside each of the $M_T$ table blocks (which are pairwise disjoint subsets of $[N]$), build a fixed causal positional $k$-ary tree of depth $L - 1$ (placing each internal node at the rightmost token of its subtree) across Layers $1, \dots, L - 1$: each internal node concatenates the raw bits of $T$ held by its at most $k$ children. Because each block has size at most $B' \le B$, every internal state fits within $B$ bits, and after Layer $L - 1$ the rightmost token of each table block holds all raw bits of $T$ in its block verbatim. Similarly, inside each of the $M_x$ disjoint address blocks in $V_x$, run a depth-$(L-1)$ causal positional $k$-ary parity-reduction tree across Layers $1, \dots, L - 1$, storing the $m$-bit vector of partial parities across $G_1, \dots, G_m$ ($m = \log_2 N \le B$ bits) at the rightmost token of each block. In Layer~$L$, token $n$ (which follows all $M_T + M_x$ designated summary tokens) uses purely positional scoring to attend to all $M_T + M_x \le k$ summary tokens, recovering both the entire truth table $T \in \{0, 1\}^N$ and the exact address $S(x) \in \{0, 1\}^m$ (by XORing the $M_x$ partial-parity vectors), and outputs $T_{S(x)}$.
\end{proof}

\section*{Acknowledgements}
Ding was supported in part by a grant from the Israel Science Foundation (ISF Grant No. 1774/20), and by the European Union (ERC, SCALE,101162665). 
Views and opinions expressed are
however those of the author(s) only and do not necessarily reflect those of the
European Union or the European Research Council. Neither the European Union
nor the granting authority can be held responsible for them.


\end{document}